\documentclass[10pt,journal,cspaper,compsoc]{IEEEtran}
\usepackage{times}
\usepackage{helvet}
\usepackage{courier}
\usepackage{amssymb}
\usepackage{amsmath}
\usepackage{enumerate}
\usepackage{epsf}
\usepackage{algorithm}
\usepackage{wrapfig}
\usepackage{sidecap}
\usepackage{color}
\usepackage{multirow}
\usepackage{rotating}
\usepackage{lscape}
\usepackage{graphicx}
\usepackage{subfigure}
\usepackage{graphicx,subfigure}
\usepackage{multicol}
\usepackage{amsmath,amsfonts,amsthm,amssymb}
\usepackage{algpseudocode}
\usepackage{footnote}
\usepackage{diagbox}

\newcommand{\data}{\mathcal{D}}
\newcommand{\ep}{\mathbb{E}}
\newcommand{\R}{\mathcal{R}}

\newcommand{\xv}{\mathbf{x}}

\newcommand{\xvtilde}{\tilde{\xv}}

\newcommand{\gv}{\mathbf{g}}

\newcommand{\Vv}{\mathbf{V}}

\newcommand{\wv}{\mathbf{w}}

\newcommand{\risk}{\mathcal{R}}

\newcommand{\muv}{\boldsymbol \mu}
\newcommand{\lambdav}{\boldsymbol \lambda }
\newcommand{\omegav}{\boldsymbol \omega }
\newcommand{\thetav}{\boldsymbol \theta} 
\newcommand{\alphav}{\boldsymbol \alpha}
\newcommand{\gammav}{\boldsymbol \gamma}

\newtheorem{lem}{Lemma}

\newcommand{\jun}[1]{{\color{blue}{\bf\sf #1}}}
\newcommand{\Ning}[1]{{\color{red}{\bf\sf [NC: #1]}}}

\begin{document}
%
\title{Dropout Training for SVMs with \\Data Augmentation}
%
%
%
%
\author{{\small Ning~Chen and
        Jun~Zhu,~\IEEEmembership{Member,~IEEE,}
        Jianfei Chen ~and~ Ting Chen}
\IEEEcompsocitemizethanks{\IEEEcompsocthanksitem N. Chen$^{\dagger}$, J. Zhu$^{\dagger}$, J. Chen$^{\ddagger}$ \ and T. Chen$^{\dagger}$ are with the MOE Key lab of Bioinformatics; Bioinformatics Division and Center for Synthetic \& Systems Biology, TNLIST; Department of Computer Science and Technology; State Key Lab of Intelligent Technology and Systems, Tsinghua University, Beijing 100084 China.\protect\\
E-mail: $^{\dagger}$\{ningchen, dcszj, tingchen\}@mail.tsinghua.edu.cn,\protect\\
 $^{\ddagger}$chenjf10@mails.tsinghua.edu.cn.\protect\\
}}

\markboth{Journal of \LaTeX\ Class Files,~Vol.~x, No.~x, May~2015}%
{Shell \MakeLowercase{\textit{et al.}}: Bare Demo of IEEEtran.cls for Computer Society Journals}
%


\IEEEcompsoctitleabstractindextext{%
\begin{abstract}
Dropout and other feature noising schemes have shown promising results in controlling over-fitting by artificially corrupting the training data. Though extensive theoretical and empirical studies have been performed for generalized linear models, little work has been done for support vector machines (SVMs), one of the most successful approaches for supervised learning. This paper presents dropout training for both linear SVMs and the nonlinear extension with latent representation learning. For linear SVMs, to deal with the intractable expectation of the non-smooth hinge loss under corrupting distributions, we develop an iteratively re-weighted least square (IRLS) algorithm by exploring data augmentation techniques. Our algorithm iteratively minimizes the expectation of a re-weighted least square problem, where the re-weights are analytically updated. For nonlinear latent SVMs, we consider learning one layer of latent representations in SVMs and extend the data augmentation technique in conjunction with first-order Taylor-expansion to deal with the intractable expected non-smooth hinge loss and the nonlinearity of latent representations. Finally, we apply the similar data augmentation ideas to develop a new IRLS algorithm for the expected logistic loss under corrupting distributions, and we further develop a non-linear extension of logistic regression by incorporating one layer of latent representations. Our algorithms offer insights on the connection and difference between the hinge loss and logistic loss in dropout training. Empirical results on several real datasets demonstrate the effectiveness of dropout training on significantly boosting the classification accuracy of both linear and nonlinear SVMs. In addition, the nonlinear SVMs further improve the prediction performance on several image datasets.
\end{abstract}

\begin{keywords}
Dropout, SVMs, logistic regression, data augmentation, iteratively reweighted least square
\end{keywords}}

\maketitle

\IEEEdisplaynotcompsoctitleabstractindextext

%
\IEEEpeerreviewmaketitle

\vspace{1cm}
\section{Introduction}\label{sec:introduction}

Artificial feature noising augments the finite training data with a large (or even infinite) number of corrupted versions, by corrupting the given training examples with a fixed noise distribution. Among the many noising schemes, dropout training~\cite{Hinton:dropout} is an effective way to control over-fitting of large deep networks by randomly omitting subsets of neurons (or features) at each iteration of a training procedure. By formulating the feature noising methods as minimizing the expectation of some loss functions under the corrupting distributions, recent work has provided theoretical understandings of such schemes from the perspective of adaptive regularization~\cite{PercyLiang13}; and has shown promising empirical results in various applications, including document classification~\cite{CorruptICML2013,PercyLiang13}, named entity recognition~\cite{Wang:emnlp13}, image classification~\cite{Wang:icml13}, tag recommendation~\cite{SDAE:AAAI2015}, etc. 

Regarding the loss functions, though much work has been done on the quadratic loss, logistic loss, or the log-loss induced from a generalized linear model (GLM)~\cite{CorruptICML2013,PercyLiang13,Wang:emnlp13}, little work has been done on the margin-based hinge loss underlying the very successful support vector machines (SVMs)~\cite{Vapnik:95}. One technical challenge is that the non-smoothness of the hinge loss makes it hard to compute or even approximate its expectation under a given corrupting distribution. Existing methods are not directly applicable, therefore calling for new solutions. This paper attempts to address this challenge and fill up the gap by extending dropout training as well as other feature noising schemes to support vector machines.

Previous efforts on learning SVMs with feature noising have been devoted to either explicit corruption or an adversarial worst-case analysis. For example, virtual support vector machines~\cite{Burges1997} explicitly augment the training data, which are usually support vectors from previous learning iterations for saving computational cost, with a finite number of additional examples that are corrupted through some invariant transformation models. A standard SVM is then learned on the corrupted data. Though simple and effective, such an approach lacks elegance and the computational cost of processing the extra corrupted examples could be prohibitive for many applications. The other work~\cite{Globerson:icml06,Dekel:icml08,Teo2008} adopts an adversarial worst-case analysis to improve the robustness of SVMs against feature deletion in testing data. Though rigorous in theory, a worst-case scenario is unlikely to be encountered in practice. Moreover, the worst-case analysis usually results in solving a complex and computationally demanding problem.


In this paper, we perform an average-case analysis and show that it is efficient to train linear SVM and nonlinear SVM predictors with latent representation learning on an infinite amount of corrupted copies of the training data by marginalizing out the corruption distributions. We concentrate on dropout training, but the results are directly extensible to other noising models, such as Gaussian, Poisson and Laplace~\cite{CorruptICML2013}. For all these noising schemes, the resulting expected hinge loss can be upper-bounded by a variational objective by introducing auxiliary variables, which follow a generalized inverse Gaussian distribution~\cite{Polson:BA11}. We apply the similar ideas on the expected logistic loss by introducing Polya-Gamma~\cite{Polson:arXiv12} distributed auxiliary variables. Specifically, we make following contributions: 

\begin{enumerate}
\item[(1)] We develop an iteratively re-weighted least square (IRLS) algorithm for dropout training of linear SVMs for both classification and regression. By minimizing a variational objective based on data augmentation, our algorithm minimizes the expectation of a re-weighted quadratic loss under the given corrupting distribution at each iteration, where the re-weights are computed in a simple closed form;
\item[(2)] We generalize the data augmentation ideas to develop an IRLS algorithm for dropout training of {\it nonlinear} SVMs that consist of one hidden layer for representation learning. 
    In order to deal with the non-smoothness of the expected hinge loss and the nonlinearity of the latent feature extractors, we apply Taylor's expansion to derive an approximate objective, and then optimize it with an iterative algorithm; 
\item[(3)] We further generalize the above ideas to develop IRLS algorithms for dropout training of logistic regression with and without one layer of nonlinear hidden units;
By sharing similar structures as those for SVMs, our IRLS algorithms shed light on the connection and difference between the hinge loss and logistic loss in the context of dropout training, complementing to the previous analysis~\cite{Rosasco:04,Globerson:07} in the supervised learning settings;
\item[(4)] We present empirical results on several image and text classification tasks and a challenging ``nightmare at test time" scenario~\cite{Globerson:icml06}. Our results demonstrate the effectiveness of our approaches, in comparison with various strong competitors.
\end{enumerate}

The rest paper is structured as follows. Section~\ref{sec:relatedwork} reviews the related work. Section \ref{sec:review} introduces the framework of learning with marginalized corrupted features. Section \ref{sec:corruptSVM} presents both linear and nonlinear dropout SVMs for classification and regression, with an iteratively re-weighted least square (IRLS) algorithm. Section \ref{sec:corruptLogist} presents both linear and nonlinear dropout logistic regression with new IRLS algorithms. Section \ref{sec:experiment} presents empirical results. Section \ref{sec:conclusion} concludes with future directions discussed.

\section{Related Work}\label{sec:relatedwork}

Dropout training has been recognized as an effective feature noising strategy for neural networks by randomly dropping hidden units during training~\cite{DropoutAlgorithm:AI2014}. One representative dropout strategy is the standard ``Monte Carlo" dropout or the explicit corruption~\cite{Dropout:2014,Srivastava:Thesis2013}, which has been applied in neural networks to prevent the feature co-adaptation effect and improve prediction performance in many applications, e.g., image classification~\cite{Hinton:dropout,Wang:icml13,Huang:TCybernetics2014}, document classification~\cite{CorruptICML2013,PercyLiang13}, named entity recognition~\cite{Wang:emnlp13}, tag recommendation~\cite{SDAE:AAAI2015}, 
online prediction with expert advice~\cite{DropoutTheory:COLT2014}, spoken language understanding~\cite{INTERSPEECH:2014}, etc. Dropout training also performs well on standard machine learning models, e.g., DART, an ensemble model of boosted regression trees using dropout training~\cite{Rashmi:2015}.

In contrast to the standard ``Monte-Carlo" dropout, in this paper, we focus on the class of models that are considered to be deterministic versions of dropout by marginalizing the noise. These models are formalized as marginalized corrupted features (MCF), and do not need the random selection. It is possible to get gradients for the marginalized loss functions. Representative work on MCF includes 
the marginalization denoising autoencoders for domain adaptation~\cite{MCFDomainAdaptation:2012} and learning nonlinear representations~\cite{DropoutAutoEncoder:2014} and marginalized dropout noise in linear regression~\cite{Srivastava:Thesis2013}. Besides, \cite{Wang:icml13} explores the idea of marginalized dropout for speed-up, and \cite{CorruptICML2013} develops several loss functions in the context of empirical risk minimization framework under different input noise distributions. Moreover, the MCF framework have also been developed for link prediction~\cite{MCFLinkPrediction:2015}, multi-label prediction~\cite{MCFSDM:2014}, image tagging~\cite{MCFImageTagging:2013} and distance metric learning~\cite{DropoutMetricLearning:2014}.

Both theoretical and empirical analyses have shown that the dropout training under MCF is equivalent to adding a regularization effect into the model for controlling over-fitting. \cite{PercyLiang13} describes how dropout can be seen as an adaptive regularizer, and \cite{PercyLiang:NIPS2014} proposes a theoretical explanation for why dropout training has been successful on high-dimensional single-layer natural language tasks. The result is that Dropout preserves the Bayes decision boundary and should therefore induce minimal bias in high dimensions. \cite{Bachman:NIPS2014} develops a pseudo-ensemble by applying dropout in perturbing the parent model and examines the relationship to the standard ensemble methods by presenting a novel regularizer based on the noising process. Other work~\cite{WhenToUseDropout:2015} analyzes some underlying problems, e.g., when the dropout-regularized criterion has a unique minimizer and when the dropout-regularization penalty goes to infinity with the weights. \cite{BayesianDropout:2015} sheds light on the dropout from a Bayesian standpoint, which enables us to optimize the dropout rates for better performance.

Though much work has been done on marginalizing the quadratic loss, logistic loss, or the log-loss induced from a generalized linear model (GLM)~\cite{CorruptICML2013,PercyLiang13,Wang:emnlp13}, little work has been done on the margin-based hinge loss underlying the very successful support vector machines (SVMs)~\cite{Vapnik:95} as discussed in Section 1. The technical challenge is that the non-smoothness of the hinge loss makes it hard to compute or even approximate its expectation under a given corrupting distribution. Existing methods are not directly applicable. This paper attempts to address this challenge and fill up the gap by extending dropout training as well as other feature noising schemes to SVMs.
Finally, some preliminary results were reported in~\cite{Chen:aaai2014} and this paper presents a systematical extension.

\section{Preliminaries}\label{sec:review}

We setup the problem in question and review the learning with marginalized corrupted features.

\subsection{Regularized loss minimization}

Consider the binary classification, where each training example is a pair $(\xv, y)$ with $\xv \in \mathbb{R}^D$ being an input feature vector and $y \in \{+1, -1\}$ being a binary label. Given a set of training data $\data = \{(\xv_n, y_n)\}_{n=1}^N$, supervised learning aims to find a function $f \in \mathcal{F}$ that maps each input to a label. To find the optimal candidate, it commonly solves a regularized loss minimization problem
\begin{eqnarray}
\min_{f \in \mathcal{F}}~ \Omega(f) + 2c \cdot \risk(\data; f),
\end{eqnarray}
where $\risk(\data; f)$ is the empirical risk of applying $f$ to the training data; $\Omega(f)$ is a regularization term to control over-fitting; and $c$ is a non-negative regularization parameter. Note that we include the factor ``$2$" simply for notation clarity as will be clear soon.

For linear models, the function $f$ is simply parameterized as $f(\xv; \wv, b) = \wv^\top \xv + b$, where $\wv$ is the weight vector and $b$ is an offset. We will denote $\thetav \triangleq\{\wv, b\}$ for clarity. Then, the regularization can be any Euclidean norms\footnote{It is a common practice to not regularize the offset.}, e.g., the $\ell_2$-norm, $\Omega(\wv) = \Vert \wv \Vert_2^2$, or the $\ell_1$-norm, $\Omega(\wv) = \Vert \wv \Vert_1$. For the loss functions, the most relevant measure is the training error, $ \sum_{n=1}^N \delta( f(\xv_n; \thetav) \neq y_n)$, which however is not easy to optimize. A convex surrogate loss is used instead, which normally upper bounds the training error. Two popular examples are the hinge loss and logistic loss\footnote{The natural logarithm is not an upper bound of the training error. We can simply change the base without affecting learning.}:
\setlength\arraycolsep{1pt} \begin{eqnarray}
\risk_h( \data; \thetav ) &=&   \sum_{n=1}^N \max\left(0, \ell - y_n f(\xv_n; \thetav)\right), \nonumber \\
\risk_l( \data; \thetav ) &=&   \sum_{n=1}^N \left(-\log p(y_n | \xv_n, \thetav) \right), \nonumber
\end{eqnarray}
\noindent where $\ell~(\geq 1)$ is the required margin, and $p(y_n | \xv_n, \thetav) \triangleq 1 / (1 + \exp(-y_n f(\xv_n; \thetav) ) )$ is the logistic likelihood. Other losses include the quadratic loss, $\sum_{n=1}^N (f(\xv_n; \thetav) - y_n)^2$, and the exponential loss, $\sum_{n=1}^N \exp( - y_n f(\xv_n; \thetav) )$, whose feature noising analyses are relatively simpler~\cite{CorruptICML2013}.

\subsection{Learning with marginalized corruption}

Let $\tilde{\xv}$ be the corrupted version of the input features $\xv$. Consider the commonly used independent corrupting model:
$$p(\tilde{\xv} | \xv )=\prod_{d=1}^D p(\tilde{x}_{d} | x_d; \eta_d),$$
where each individual distribution is a member of the exponential family, with the natural parameter $\eta_d$. Another common assumption is that the corrupting distribution is unbiased, that is, $\ep_p[\tilde{\xv}  | \xv ] = \xv$, where we use $\ep_p[\cdot] \triangleq \ep_{p(\tilde{\xv}|\xv)}[\cdot]$ to denote the expectation taken over the corrupting distribution $p(\tilde{\xv} | \xv)$. Such examples include the unbiased blankout (or dropout) noise, Gaussian noise, Laplace noise, and Poisson noise~\cite{Vincent:2008,CorruptICML2013}.

For the {\it explicit corruption}~\cite{Burges1997}, each example $(\xv_n, y_n)$ is corrupted $M$ times from the corrupting model $p(\tilde{\xv}_n|\xv_n)$, resulting in the corrupted examples $(\tilde{\xv}_{nm}, y_n)$, $m \in [M]$. This procedure generates a new corrupted dataset $\tilde{\data}$ with a larger size of $NM$. Then, the model can be learned on the generated dataset by minimizing the average loss function over $M$ corrupted data points:
\begin{eqnarray}\label{eq:ExplicitCorruption}
\mathcal{L}(\tilde{\data}; \thetav)=\sum_{n=1}^N \frac{1}{M} \sum_{m=1}^{M}\mathcal{R}(\tilde{\xv}_{nm}, y_n; \thetav),
\end{eqnarray}
\noindent where $\mathcal{R}(\xv, y; \thetav)$ is the loss function of the model incurred on the training example $(\xv, y)$. As $\mathcal{L}(\tilde{\data}; \thetav)$ scales linearly with the number of corrupted observations, this approach may suffer from a high computational cost whenever $M$ is moderately large.

Dropout training adopts the strategy of {\it implicit corruption}, which learns the model with marginalized corrupted features by minimizing the expectation of a loss function under the corrupting distribution
\begin{eqnarray}\label{eq:ExpectedLoss}
\mathcal{L}(\data; \thetav) = \sum_{n=1}^{N}\ep_{p}[ \R(\tilde{\xv}_n, y_n; \thetav)].
\end{eqnarray}
\noindent The objective can be seen as a limit case of~(\ref{eq:ExplicitCorruption}) when $M \to \infty$, by the law of large numbers. Such an expectation scheme has been widely adopted~\cite{PercyLiang13,CorruptICML2013,Wang:emnlp13,Wang:icml13}.

The choice of the loss function $\mathcal{R}$ in (\ref{eq:ExpectedLoss}) can make a significant difference, in terms of computation cost and prediction accuracy. Previous work on feature noising has covered the quadratic loss, exponential loss, logistic loss, and the loss induced from generalized linear models (GLM). For the quadratic loss and exponential loss, the expectation in Eq.~(\ref{eq:ExpectedLoss}) can be computed analytically, thereby leading to simple gradient descent algorithms~\cite{CorruptICML2013}. However, it does not have a closed form to compute the expectation for the logistic loss or the GLM log-loss. Previous analysis has resorted to approximation methods, such as using the second-order Taylor expansion~\cite{PercyLiang13} or an upper bound by applying Jensen's inequality~\cite{CorruptICML2013}, both of which lead to effective algorithms in practice. In contrast, little work has been done on the hinge loss, for which the expectation under corrupting distributions cannot be analytically computed either, therefore calling for new algorithms.


\vspace{-.05cm}
\section{Dropout Support Vector Machines}\label{sec:corruptSVM}
\vspace{-.05cm}

In this section, we present dropout training for both linear SVMs and its nonlinear extension with representation learning in the context of classification and regression.

\vspace{-.05cm}
\subsection{Linear SVMs with Corrupting Noise}
\vspace{-.05cm}

For linear SVMs, the expected hinge loss can be written as
\begin{eqnarray}\label{eqn:EpHingeLoss}
\mathcal{L}_h(\data; \thetav) = \sum_{n=1}^N \ep_{p}[\max{(0, \zeta_n)}],
\end{eqnarray}
where we define $\zeta_n \triangleq  \ell - y_n(\wv^\top\tilde{\xv}_n)$.\footnote{We treat the offset $b$ implicitly by augmenting $\xv_n$ and $\tilde{\xv}_n$ with one dimension of deterministic $1$. More details will be given in the algorithm.}
Following the regularized loss minimization framework, we define the optimization problem of SVMs with marginalized corrupted features as
\begin{eqnarray}\label{eq:drop-svm}
\min_{\thetav} \Vert \wv \Vert_2^2 + 2c \cdot \mathcal{L}_h(\data; \thetav).
\end{eqnarray}
Below, we present a simple iteratively re-weighted least square (IRLS) algorithm to solve this problem. 
Our method consists of a variational bound of the expected loss and a simple algorithm that iteratively minimizes an expectation of a re-weighted quadratic loss. We also apply the similar ideas to develop a simple IRLS algorithm for minimizing the expected logistic loss in Section~\ref{sec:corruptLogist}, thereby allowing for a systematical comparison of the hinge loss with the logistic and quadratic losses in the context of feature noising.

\subsubsection{A variational bound with data augmentation}

Since we do not have a closed-form expression of the expectation of the max function, it is intractable to directly solve problem~(\ref{eq:drop-svm}). Here, we derive a variational upper bound based on a data augmentation formulation of the expected hinge loss. Specifically, let $\phi(y_n|\tilde{\xv}_n, \thetav) = \exp\{-2c \max(0, \zeta_n)\}$  be the unnormalized pseudo-likelihood\footnote{Pseudo-likelihood has been widely used in statistics. Here, we simply mean that the likelihood is not well normalized.} of the response variable for sample $n$. Then we have
\begin{eqnarray}
2c \cdot \mathcal{L}_h(\data; \thetav) = - \sum_n \ep_p [ \log \phi(y_n | \tilde{\xv}_n, \thetav) ].
\end{eqnarray}
Using the ideas of data augmentation~\cite{Polson:BA11,Zhu:jmlr14}, the pseudo-likelihood can be expressed as
\setlength\arraycolsep{-3pt} \begin{eqnarray}\label{eqn:lemmaScaleMixture}
&& \phi(y_n | \tilde{\xv}_n, \thetav) = \int_0^\infty \!\!\! \frac{1}{\sqrt{2\pi\lambda_n}}\exp\left\{-\frac{(\lambda_n+c\zeta_n)^2}{2\lambda_n}\right\}d\lambda_n,
\end{eqnarray}
where $\lambda_n$ is the augmented variable associated with data $n$. Using (\ref{eqn:lemmaScaleMixture}) and Jensen's inequality,
we can derive a variational upper bound of the expected hinge loss multiplied by the factor $2c$ (i.e., $2c \cdot \mathcal{L}_h(\data; \thetav)$) as
\setlength\arraycolsep{1pt} \begin{eqnarray}\label{eqn:Upperbound}
\mathcal{L}_h(\thetav, q(\lambdav)) &=& -H(\lambdav) + \sum_{n=1}^N \Big\{ \frac{1}{2}\ep_q \lbrack \log\lambda_n\rbrack\\
&& + \ep_q \Big\lbrack\frac{1}{2\lambda_n}\ep_{p} (\lambda_n + c\zeta_n)^2\Big\rbrack \Big\} + c^\prime, \nonumber
\end{eqnarray}
where $H(\lambdav)$ is the entropy of the variational distribution $q(\lambdav)$ with $\lambdav \triangleq \{\lambda_n\}_{n=1}^N$; 
$c^\prime$ is a constant; and we have defined $\ep_q[\cdot] \triangleq \ep_{q(\lambdav)}[\cdot]$ to denote the expectation taken over a variational distribution $q$. Now, our variational optimization problem is
\begin{eqnarray}\label{eq:variationalBound}
\min_{\thetav, q(\lambdav) \in \mathcal{P} } \Vert \wv \Vert_2^2 + \mathcal{L}_h(\thetav, q(\lambdav) ),
\end{eqnarray}
where $\mathcal{P}$ is the simplex space of normalized distributions. We should note that when there is no feature noise (i.e., $\tilde{\xv} = \xv$), the bound is tight and we are learning the standard SVM classifier. Please see Appendix~A for the derivation. We will empirically compare with SVM in experiments.

\subsubsection{Iteratively Re-weighted Least Square Algorithm}

In the upper bound, we note that when the variational distribution $q(\lambdav)$ is given, the term $\ep_{p} [(\lambda_n + c\zeta_n)^2]$ is an expectation of a quadratic loss, which can be analytically computed. 
We leverage this property and develop a coordinate descent algorithm to solve problem~(\ref{eq:variationalBound}). Our algorithm iteratively solves the following two steps, analogous to the common procedure of a variational EM algorithm.

{\bf For $q(\lambdav)$ (i.e., E-step)}: when the parameters $\thetav$ are given, this step involves inferring the variational distribution $q(\lambdav)$. Specifically, optimizing $\mathcal{L}$ over $q(\lambdav)$, we get $q(\lambdav) = \prod_n q(\lambda_n)$ and each term is:
\setlength\arraycolsep{1pt} \begin{eqnarray}\label{eq:svm-qlambda}
    q(\lambda_n) &\propto& \frac{1}{\sqrt{\lambda_n}}\exp\left\{-\frac{1}{2}\left( \lambda_n + \frac{c^2\ep_p [\zeta_n^2]}{\lambda_n} \right)\right\}\nonumber\\
    &=&\mathcal{GIG}\left( \lambda_n; \frac{1}{2}, 1, c^2\ep_{p}[\zeta_n^2] \right),
    \end{eqnarray}
    where the second-order expectation is
    \begin{eqnarray}
    \ep_p[\zeta_n^2] =&&  \wv^\top(\ep_p[\tilde{\xv}_n]\ep_p[\tilde{\xv}_n ]^\top  + V_p[\tilde{\xv}_n]) \wv \nonumber \\
    && - 2 \ell  y_n \wv^\top\ep_p[\tilde{\xv}_n] + \ell^2;
    \end{eqnarray}
    and $V_p[\tilde{\xv}_n]$ is a $D\times D$ diagonal matrix with the $d$th diagonal element being the variance of $\tilde{x}_{nd}$, under the corrupting distribution $p(\tilde{\xv}_n|\xv_n)$. We have denoted $\mathcal{GIG}(x;p,a,b) \propto x^{p-1}\exp(-\frac{1}{2}(\frac{b}{x}+ax))$ as a generalized inverse Gaussian distribution~\cite{Devroye:book1986}. Thus, $\lambda_n^{-1}$ follows an inverse Gaussian distribution
    \begin{eqnarray}\label{eqn:inverseGaussian}
    q(\lambda_n^{-1}|\tilde{\xv}_n, \thetav) = \mathcal{IG}\left( \lambda_n^{-1}; \frac{1}{c\sqrt{\ep_p[\zeta_n^2]}}, 1 \right).
    \end{eqnarray}

{\bf For $\thetav \triangleq \wv$ (i.e., M-step)}: by fixing $q(\lambdav)$ and removing irrelevant terms, this substep minimizes the objective:
    \setlength\arraycolsep{-2pt} \begin{eqnarray}\label{eqn:Upperbound-w-b}
    && \mathcal{L}_{[\thetav]} 
                           = \Vert \wv \Vert_2^2  + \sum_{n=1}^N  \ep_{p}\left[c \zeta_n  +  \frac{c^2}{2} \gamma_n \zeta_n^2 \right],
    \end{eqnarray}
    where $\gamma_n \triangleq \ep_q[\lambda_n^{-1}]$. We observe that this substep is equivalent to minimizing the expectation of a re-weighted quadratic loss, as summarized in Lemma 1, whose proof is deferred to Appendix B for brevity.
\begin{lem}\label{lemma:CorruptSVM}
Given $q(\lambdav)$, the M-step minimizes the re-weighted quadratic loss (with the $\ell_2$-norm regularizer):
\begin{eqnarray}
\Vert \wv \Vert_2^2 + \frac{c^2}{2} \sum_n \gamma_n \ep_{p} \left[ (\wv^\top \tilde{\xv}_n - y_n^h)^2 \right],
\end{eqnarray}
where $y_n^h = (\ell + \frac{1}{c \gamma_n}) y_n$ is the re-weighted label, and the re-weights are computed in closed-form:
\begin{eqnarray}\label{eq:ExpLambda}
\gamma_n \triangleq \ep_{q}[\lambda_n^{-1}] = \frac{1}{c\sqrt{\ep_p[\zeta_n^2]}}.
\end{eqnarray}
\end{lem}
For low-dimensional data, we can do matrix inversion to get the closed-form solution
\footnote{To consider offset, we simply augment $\xv$ and $\tilde{\xv}$ with an additional unit of $1$. The variance $V_p[\tilde{\xv}_n]$ is augmented accordingly. The identity matrix $I$ is augmented by adding one zero row and one zero column.}:
\setlength\arraycolsep{-3pt} \begin{eqnarray}\label{eq:drop-svm-analytical-solution}
&&\wv \!= \!\!\left(\! \frac{2}{c^2} I +  \sum_{n=1}^N \! \gamma_n \ep_p[ \tilde{\xv}_n \tilde{\xv}_n^\top] \!\right)^{-1} \!\!\! \left(\!\sum_{n=1}^N \! \gamma_n y_n^h \ep_p[\tilde{\xv}_n] \! \right),
\end{eqnarray}
where $\ep_p[ \tilde{\xv}_n \tilde{\xv}_n^\top ] =  \ep_p[\tilde{\xv}_n] \ep_p[\tilde{\xv}_n]^\top + V_p[\tilde{\xv}_n]$. However, if the data lies in a high-dimensional space, e.g., text documents in a bag-of-words vector space with tens of thousands of dimensions, the above matrix inversion will be computationally expensive. In such cases, we can use numerical methods, e.g., the quasi-Newton method~\cite{Liu:89}, to efficiently solves for $\thetav$.

To summarize, our algorithm iteratively minimizes the expectation of a simple re-weighted quadratic loss under the given corrupting distribution, where the re-weights $\gamma_n$ are computed in an analytical form. Therefore, it is an extension of the classical {\it iteratively re-weighted least square} (IRLS) algorithm~\cite{Hastie:2009} in order to deal with dropout training. 
We also observe that if we fix $\gamma_n$ at $\frac{1}{c}$ and set $\ell=0$, we are minimizing the quadratic loss under the corrupting distribution, as studied in~\cite{CorruptICML2013}.
We will empirically show that our iterative algorithm for the expected hinge-loss consistently improves over the standard quadratic loss by adaptively updating $\gamma_n$.
Finally, as we assume that the corrupting distribution is unbiased, i.e., $\ep_{p}[\tilde{x}_{nd} | x_{nd} ] = x_{nd}$, we only need to compute the variance of the corrupting distribution, i.e., $V_p[\tilde{x}_{nd}]=\frac{q}{1-q}x_{nd}^2$ for dropout distribution, which is easy for all the existing exponential family distributions. An overview of the variance of the commonly used corrupting distributions can be found in \cite{CorruptICML2013}.

\subsection{Dropout SVMs with Representation Learning}\label{sec:NLSVM}

We have assumed that the classifier is a linear model with respect to the input features. This assumption can be relaxed by learning a nonlinear representation, as popularized in representation learning~\cite{Bengio:PAMI13}. Here we present an extension to learn a nonlinear mapping of the input features.

Let $\gv(\xv;\alphav)$ denote a $K$-dimensional nonlinear transformation of the $D$-dimensional input features $\xv$, parameterized by a $D\times K$ matrix $\alphav$. For example, we can define the logistic transformation, each element $k$ is
$$g_k(\xv; \alphav) = \textrm{Sigmoid}( \alphav_k^\top \xv ) \triangleq \frac{1}{ 1 + \exp(- \alphav_k^\top \xv ) },$$
where $\alphav_k$ is the $k$th column of $\alphav$. We then define our linear discriminant function\footnote{The offset is again ignored for simplicity.} as $$f(\xv; \wv, \alphav) = \wv^\top \gv(\xv; \alphav)$$
where $\wv \in \mathbb{R}^K$ is the vector of classifier weights. We still let $\zeta_n = \ell - y_n (\wv^\top \gv(\xv;\alphav))$. Then, we have the same expected hinge loss as in Eq.~(\ref{eqn:EpHingeLoss}).

Using the same data augmentation technique, we can derive a variational upper bound of the expected hinge loss as in Eq.~(\ref{eqn:Upperbound}), again with the new definition of $\zeta_n$. However, note that the variational bound is also a function of $\alphav$. With the nonlinear transformation, the challenge is on computing the variational bound, which is intractable in general. Here, we apply the Taylor-expansion of $\gv(\cdot)$ in order to get an approximation. Specifically, we have
\setlength\arraycolsep{-4pt} \begin{eqnarray}\label{eqn:taylorExpension}
&& \gv(\xvtilde_n) \approx \gv^\prime(\xvtilde_n) \triangleq \gv(\muv_n) \! + \! \nabla_{\xvtilde} \gv(\muv_n)^\top \! (\xvtilde_n - \muv_n),
\end{eqnarray}
where $\nabla_{\xvtilde} \gv(\muv_n)$ is the first-order derivative of $\gv(\xvtilde,\alphav)$ with respect to $\xvtilde$ evaluated at $\muv_n$, a $D\times K$ matrix with each element being $\nabla_{\xvtilde} g_{dk}(\muv_n) = g_k(\muv_n)(1-g_k(\muv_n))\alpha_{dk}$; and $\muv_n = \ep_p[\xvtilde_n]$ is the mean of the corrupted features. For unbiased corrupting noise, we have $\muv_n = \xv_n$.

With the first-order Taylor expansion, we can compute the variational bound, which involves the variance of the corruption. Then, an alternating minimization algorithm can be developed to iteratively update the following steps: 

{\bf For $q(\lambdav)$}: the solution has the same form as in (\ref{eq:svm-qlambda}):
\setlength\arraycolsep{1pt} \begin{eqnarray}
    q(\lambda_n) 
    &=&\mathcal{GIG}\left( \lambda_n; \frac{1}{2}, 1, c^2\ep_{p}[\zeta_n^2] \right),\nonumber
\end{eqnarray}
but with the new definition of $\zeta_n$. Under the Taylor-expansion, we have the second-order expectation:
\begin{eqnarray}
    \ep_p[\zeta_n^2] =&&  \wv^\top(\ep_p[\gv^\prime(\tilde{\xv}_n) \gv^\prime(\tilde{\xv}_n)^\top]) \wv \nonumber \\
    && - 2 \ell  y_n \wv^\top\ep_p[\gv^\prime(\tilde{\xv}_n)] + \ell^2\nonumber 
\end{eqnarray}
where $\ep_p[\gv^\prime(\tilde{\xv}_n)] = \gv(\muv_n)$, $\ep_p[\gv^\prime(\tilde{\xv}_n) \gv^\prime(\tilde{\xv}_n)^\top] = \gv(\muv_n)\gv(\muv_n)^\top + \triangledown_{\xvtilde} \gv(\muv_n)^\top V_p[\tilde{\xv}_{n}]\triangledown_{\xvtilde} \gv(\muv_n)$, and $V_p[\tilde{\xv}_n]$ is again a $D\times D$ diagonal matrix with the $d$th diagonal element being the variance of $\tilde{x}_{nd}$, under the corrupting distribution $p(\tilde{\xv}_n|\xv_n)$. 


{\bf For $\wv$}: this step is similar as in the linear case, but with subtle change on the features. For ease of computation, we keep the objective that only includes $\wv$:
\begin{eqnarray}\label{eq:svm-w-subproblem}
\mathcal{L}_{[\wv]} &=& \Vert \wv \Vert_2^2  + \sum_{n=1}^N  \ep_{p}\left[c \zeta_n  +  \frac{c^2}{2} \gamma_n \zeta_n^2 \right],
\end{eqnarray}
where the coefficient $\gamma_n$ is computed as:
\begin{eqnarray}\label{eqn:NLSVM_gamma}
\gamma_n = \ep_q[\lambda_n^{-1}]=\frac{1}{c\sqrt{\ep_p[\zeta_n^2]}}.
\end{eqnarray}
If the number of latent features is not large (i.e., $K$ is small), we can get the optimal solution in the same closed-form as (\ref{eq:drop-svm-analytical-solution}), by simply replacing $\xvtilde_n$ with $\gv^\prime(\tilde{\xv}_n)$.
However, if $K$ is large, we must resort to numerical methods (e.g., quasi-Newton methods) and use the derivative for $\wv$:
\begin{eqnarray}
\frac{\partial \mathcal{L}_{[\wv]}}{\partial w_k} &=& 2 w_k + c^2 \sum_n \gamma_n \!\Big((\tilde{y}_n\!-\!y_n^h) g_k(\muv_n) \nonumber \\
&& + \sum_d V_p[\tilde{x}_{nd}]\tilde{h}_{nd}\triangledown_{\tilde{\xv}} g_{dk}(\muv_n)\Big),\nonumber
\end{eqnarray}
where $\tilde{y}_n=\sum_k w_k g_k(\muv_n)$, $y_n^h = (\ell + \frac{1}{c \gamma_n}) y_n$ and $\tilde{h}_{nd} = \sum_k w_k \nabla_{\xvtilde} g_{dk}(\muv_n)$.

{\bf For $\alphav$}: this is the new step, which can be done by gradient descent. The objective including all the $\alphav$ is
\begin{eqnarray}\label{eq:svm-alpha-subproblem}
\mathcal{L}_{[\alphav]} = \|\alphav\|_2^2 
 + \sum_{n=1}^N  \ep_{p}\left[c \zeta_n  +  \frac{c^2}{2} \gamma_n \zeta_n^2 \right],
\end{eqnarray}
and now $\alphav$ could be solved using gradient descent. The gradient for $\alpha_{dk}$ is
\begin{eqnarray}
\frac{\partial \mathcal{L}_{[\alphav]}}{\partial \alpha_{dk}} &=& 2 \alpha_{dk}
+ c^2 \sum_n \gamma_n \Big((\tilde{y}_n - y_n^h)\mu_{nd} \nonumber \\
&& + V_p[\tilde{x}_{nd}]\tilde{h}_{nd}\rho_{kd}\Big)w_k \eta_k,\nonumber
\end{eqnarray}
where the coefficients are computed as $\rho_{kd}=(1+(1-2g_k(\muv_n))\alpha_{dk}\mu_{nd})$ and $\eta_k = g_k(\muv_n) (1- g_k(\muv_n))$.

In summary, the nonlinear SVMs can be learned via coordinate descent by iteratively updating $\gammav$ with Eq. (\ref{eqn:NLSVM_gamma}), solving for the model parameters $\wv$ by minimizing $\mathcal{L}_{[\wv]}$ in (\ref{eq:svm-w-subproblem}), and solving for the transformation weights $\alphav$ by minimizing $\mathcal{L}_{[\alphav]}$ in (\ref{eq:svm-alpha-subproblem}).

\subsection{Dropout SVMs for Regression}

We briefly discuss how to extend the above ideas to the regression task, where the response variable $Y$ takes real values. For regression, a widely used loss for support vector regression (SVR) models~\cite{Smola:03} is the
$\epsilon$-insensitive loss:
\begin{eqnarray}
\mathcal{R}_\epsilon(\data; \thetav)=\sum_{n=1}^N \max(0, |\Delta_n|-\epsilon),
\end{eqnarray}
where $\Delta_n \triangleq y_n - \wv^\top\xv_n$ is the difference between the true value and the model prediction, and $\epsilon$ is a pre-defined positive constant.
For dropout training, we consider the expected loss $\mathcal{L}_\epsilon = \ep_{p} [\mathcal{R}_\epsilon]$.
To deal with the intractability, we develop a similar IRLS algorithm with data augmentation.
Specifically, let $\varphi(y_n|\tilde{\xv}_n, \thetav)=\exp\{-2c \max(0, |\Delta_n|-\epsilon)\}$ be the pseudo-likelihood of the response variable for sample $n$. We have
\begin{eqnarray}
2c \cdot \mathcal{L}_\epsilon(\data; \thetav) = - \sum_n \ep_p [ \log \varphi(y_n | \tilde{\xv}_n, \thetav) ].
\end{eqnarray}
By noting the equality that $\max(0, |\Delta_n|-\epsilon)= \max(0, \Delta_n-\epsilon) + \max(0, -\Delta_n-\epsilon)$ and applying the
ideas of data augmentation in Eq. (\ref{eqn:lemmaScaleMixture}), we have: 
\begin{eqnarray}\label{eqn:Upperbound-svr}
\varphi(y_n|\tilde{\xv}_n, \thetav) &=& \int_{0}^\infty\frac{1}{\sqrt{2\pi\lambda_n}}\exp\left\{-\frac{\lambda_n+c(\Delta_n-\epsilon)^2}{2\lambda_n}\right\}d\lambda_n\nonumber\\
&\times& \int_{0}^\infty\frac{1}{\sqrt{2\pi\omega_n}}\exp\left\{-\frac{\omega_n-c(\Delta_n+\epsilon)^2}{2\omega_n}\right\}d\omega_n , \nonumber
\end{eqnarray}
where $(\lambda_n, \omega_n)$ are a pair of augmented variables associated with data $n$.
Then, using this data augmentation expression and 
Jensen's inequality, we can derive a variational upper bound $\mathcal{L}_\epsilon$ of the expected $\epsilon$-insensitive loss again multiplied by the factor $2c$ (i.e., $2c \cdot \mathcal{L}_\epsilon(\data; \thetav)$) as
\setlength\arraycolsep{-1pt}\begin{eqnarray}
&&\mathcal{L}_\epsilon(\thetav; q(\lambdav,\omegav)) = - H(\lambdav, \omegav) + \sum_{n=1}^N  \Big\{\frac{1}{2}\ep_q[\log \lambda_n + \log \omega_n ] \nonumber\\
&&+\ep_q \Big[ \frac{\ep_p(\lambda_n+c(\Delta_d-\epsilon))^2}{2\lambda_n} + \frac{\ep_p(\omega_n-c(\Delta_d+\epsilon))^2}{2\omega_n} \Big] \Big\} + c^\prime \nonumber
\end{eqnarray}
where $H(\lambdav,\omegav)$ is the entropy of the variational distribution $q(\lambdav, \omegav)$. 
Then the optimization problem for marginalized corrupted SVR is
\setlength\arraycolsep{1pt}\begin{eqnarray}\label{eq:drop-svr}
\min_{\thetav, q(\lambdav, \omegav)\in \mathcal{P}}~~\|\wv\|_2^2 + \mathcal{L}_\epsilon(\thetav; q(\lambdav,\omegav)).
\end{eqnarray}

In the upper bound, we note that when the variational distribution $q(\lambdav, \omegav)$ is given, the term $\ep_p[(\lambda_n+c(\Delta_n-\epsilon))^2]$ is an expectation of a quadratic loss, which can be analytically computed. Similar as the classification case, we develop an IRLS algorithm for problem~(\ref{eq:drop-svr}) with the following two steps.

{\bf For $q(\lambdav, \omegav)$ (i.e., E-step)}: infer the variational distribution $q(\lambdav, \omegav)$. Optimize $\mathcal{L}$ over $q(\lambdav, \omegav)$, we get $q(\lambdav, \omegav) = \prod_n q(\lambda_n) q(\omega_n)$ and each term is:
\begin{eqnarray}
q(\lambda_n) 
&=&\mathcal{GIG}\left(\lambda_n; \frac{1}{2}, 1, c^2\ep_p[(\Delta_n-\epsilon)^2]\right), \nonumber \\
q(\omega_n) 
&=&\mathcal{GIG}\left(\omega_n; \frac{1}{2}, 1, c^2\ep_p[(\Delta_n + \epsilon)^2]\right), \nonumber
\end{eqnarray}
where the second-order expectations are
$\ep_p[(\Delta_n-\epsilon)^2] = \wv^\top(\ep_p[\tilde{\xv}_n]\ep_p[\tilde{\xv}_n ]^\top  + V_p[\tilde{\xv}_n]) \wv - 2(y_n-\epsilon) \wv^\top\ep_p[\tilde{\xv}_n] + (y_n-\epsilon)^2$
and $\ep_p[(\Delta_n+\epsilon)^2]=\wv^\top(\ep_p[\tilde{\xv}_n]\ep_p[\tilde{\xv}_n ]^\top+V_p[\tilde{\xv}_n]) \wv -2(y_n+\epsilon) \wv^\top\ep_p[\tilde{\xv}_n] + (y_n+\epsilon)^2$.
Thus, $\lambda_n^{-1}$ and $\omega_n^{-1}$ follow inverse Gaussian distributions:
\begin{eqnarray}
q(\lambda_n^{-1}|\tilde{\xv}_n, \theta)& = & \mathcal{IG}\left(\lambda_n^{-1}; \frac{1}{c\sqrt{\ep_p[(\Delta_n-\epsilon)^2]}}, 1\right) \\
q(\omega_n^{-1}|\tilde{\xv}_n, \theta) & = & \mathcal{IG}\left(\omega_n^{-1}; \frac{1}{c\sqrt{\ep_p[(\Delta_n+\epsilon)^2]}}, 1\right). 
\end{eqnarray}

{\bf For $\thetav\triangleq\wv$ (i.e., M-step)}: removing the irrelevant terms, this step involves minimizing the following objective:
\begin{eqnarray}
\mathcal{L}_{[\thetav]} = \|\wv\|_2^2 + \frac{c^2}{2}\sum_{n=1}^N\ep_p\left[\gamma_n(\Delta_n-\epsilon)^2+\delta_n(\Delta_n+\epsilon)^2\right], \nonumber
\end{eqnarray}
where $\gamma_n \triangleq \ep_q[\lambda_n^{-1}]$ and $\delta_n \triangleq \ep_q[\omega_n^{-1}]$. Similar as in Lemma \ref{lemma:CorruptSVM}, it can be observed that this substep can also be equivalent to solving a re-weighted quadratic loss, as summarized in Lemma \ref{lemma:SVR}.
\begin{lem}\label{lemma:SVR}
Given $q(\lambdav, \omegav)$, the M-step minimizes the re-weighted quadratic loss (with the $\ell_2$-norm regularizer):
\begin{eqnarray}
\Vert \wv \Vert_2^2 + \frac{c^2}{2} \sum_n (\gamma_n+\delta_n) \ep_{p} \left[ (\wv^\top \tilde{\xv}_n - y_n^\epsilon )^2 \right],
\end{eqnarray}
where $y_n^\epsilon \triangleq (y_n+\frac{\delta_n-\gamma_n}{\delta_n+\gamma_n}\epsilon)$ is the re-weighted response, and the re-weights are computed in closed-form:
\begin{eqnarray}
\gamma_n = \frac{1}{c\sqrt{\ep_p[(\Delta_n-\epsilon)^2]}}, ~ \delta_n = \frac{1}{c\sqrt{\ep_p[(\Delta_n+\epsilon)^2]}}.\nonumber
\end{eqnarray}
\end{lem}
Similar as in the classification case, we can solve for the closed-form solution by using matrix inversion for low-dimensional data, while for high-dimensional data, we must resort to numerical approaches. 

\section{Dropout Logistic Regression}\label{sec:corruptLogist}

In this section, we develop a new IRLS algorithm for dropout training of logistic regression and its extension to learn latent representations for classification.
Our IRLS algorithm also iteratively minimizes the expectation of a re-weighted quadratic loss under the corrupting distribution and computes the re-weights analytically. Such an IRLS algorithm allows us to draw comparisons with SVMs.

\subsection{Logistic Regression with Corrupting Noise}

Define $\omega_n \triangleq \wv^\top \tilde{\xv}_n$.
The expected logistic loss under a corrupting distribution is
\begin{eqnarray}\label{eqn:EpLogistLoss}
\mathcal{L}_l (\data; \wv) =  -\sum_{n=1}^N \ep_{p}\left[\log\left(\frac{e^{y_n \omega_n}}{1+e^{y_n\omega_n}}\right)\right].
\end{eqnarray}
Again since the expectation cannot be computed in closed-form, we derive a variational bound as a surrogate.
Specifically, let $\psi(y_n |\tilde{\xv}_n, \wv) 
=\frac{e^{c y_n\omega_n}}{(1+e^{y_n\omega_n})^c}$ be the pseudo-likelihood of the response variable for sample $n$.
We have $c \cdot \mathcal{L}_l(\data; \wv) = - \sum_n \ep_p[ \log \psi(y_n | \tilde{\xv}_n, \wv) ]$.\footnote{We drop the constant factor $2$ again for notation simplicity.}
Using data augmentation techniques~\cite{Polson:arXiv12,Chen:IJCAI2013}, the pseudo-likelihood can be expressed as
\setlength\arraycolsep{-4pt} \begin{eqnarray}\label{eqn:PolyaGammaEquality}
&& \psi(y_n | \tilde{\xv}_n, \wv)  = \frac{1}{2^c} e^{\kappa_n \omega_n}\int_0^\infty e^{-\frac{\lambda_n(y_n\omega_n)^2}{2}}p(\lambda_n)d\lambda_n,
\end{eqnarray}
where $\kappa_n \triangleq \frac{c}{2}y_n$ and $\lambda_n$ is the augmented Polya-Gamma variable following distribution $p(\lambda_n)\sim \mathcal{PG}(\lambda_n; c, 0)$. Using (\ref{eqn:PolyaGammaEquality}), we can derive the upper bound of the expected logistic loss multiplied by the factor $c$ (i.e., $c \cdot \mathcal{L}_l(\data; \wv)$):
\setlength\arraycolsep{1pt} \begin{eqnarray}\label{eqn:LogisticRegression}
\mathcal{L}_l^\prime(\wv, q(\lambdav)) &=& - H(\lambdav) + \sum_{n=1}^N \Big\{\frac{1}{2}\ep_q[\lambda_n]\ep_p[\omega_n^2]  \\
&& -\ep_q[\log p(\lambda_n)] -\frac{c}{2}y_n \ep_p[\omega_n ] \Big\} + c^\prime,\nonumber
\end{eqnarray}
and get the variational optimization problem
\begin{eqnarray}
\min_{\wv, q(\lambdav) \in \mathcal{P}} \Vert\wv\Vert^2_2 + \mathcal{L}_l^\prime(\wv, q(\lambdav)),
\end{eqnarray}
where $q(\lambdav)$ is the variational distribution


\begin{table*}[t]\vspace{-.2cm}
\caption{Comparison of hinge loss, logistic loss and quadratic loss under the IRLS algorithmic framework.}\label{table:lossComparison}\vspace{-.4cm}
\begin{center}
 \scalebox{1}
 { \setlength{\tabcolsep}{1.8pt}
       \begin{tabular}{|c|c|c|c|c|c|}
       \hline
        \hline
        \multirow{2}{*}{\diagbox{Types}{Settings}} & \multicolumn{2}{c|}{Hyper-parameters} & \multicolumn{2}{c|}{Re-weights Update} & \multirow{2}{*}{Reduction to Quadratic Loss} \\
        \cline{2-5}
           {}         & parameter $\ell$ & parameter $c$ & update $\gamma_n$ & update $y_n$ & {} \\
        \hline
        Hinge-loss       & $\ell$  & $c$ & Eq.~(\ref{eq:ExpLambda}) & $y_n^h$ & $\ell=0,~ \gamma_n = \frac{1}{c}$ \\
        Logistic-loss    & --  & $c$ & Eq.~(\ref{eq:ExpLambda2}) &  $y_n^l$ & $\gamma_n = \frac{c}{2}$ \\
        \hline
        \end{tabular}
}
\end{center}\vspace{-.5cm}
\end{table*}

We solve the variational problem with a coordinate descent algorithm as follows:

{\bf For $q(\lambdav)$ (i.e., E-step)}: optimizing $\mathcal{L}^\prime$ over $q(\lambdav)$, we have $q(\lambdav) = \prod_n q(\lambda_n)$ and each term is:
\begin{eqnarray}\label{eq:lr-qlambda}
q(\lambda_n) &\propto& \exp\left(-\frac{1}{2}\lambda_n \ep_p[\omega_n^2]\right)p(\lambda_n | c,0)\nonumber\\
& = &\mathcal{PG}\left( \lambda_n; c, \sqrt{\ep_p[\omega_n^2]} \right),
\end{eqnarray}
which is a Polya-Gamma distribution with $\ep_p[\omega_n^2]=\wv^\top(\ep_p[\tilde{\xv}_n]\ep_p[\tilde{\xv}_n]^\top + V_p[\tilde{\xv}_n])\wv$. 

{\bf For $\wv$ (i.e., M-step)}: by fixing $q(\lambdav)$ and removing irrelevant terms, this step minimizes the objective
\setlength\arraycolsep{-3pt} \begin{eqnarray}
&& \mathcal{L}_{[\wv]}^\prime = \Vert\wv\Vert_2^2 + \sum_{n=1}^N \frac{1}{2}\ep_q[\lambda_n] \ep_p[\omega_n^2] - \frac{c}{2}y_n\ep_p[\omega_n].
\end{eqnarray}
If the data is not in a high dimensional space, we can get the optimal solution in a closed-form\footnote{The offset can be similarly incorporated as in the hinge loss.}:
\setlength\arraycolsep{-3pt} \begin{eqnarray}\label{eq:logist-analytical-solution}
&&\wv = \left(\! I + \frac{1}{2} \!\sum_{n=1}^N \! \ep_q[\lambda_n] \ep_p[ \tilde{\xv}_n \tilde{\xv}_n^\top ] \! \right)^{-1} \!\!\!\!\! \left(\!\frac{c}{4} \!\! \sum_{n=1}^N \! y_n \ep_p[\tilde{\xv}_n] \!\right)\!.
\end{eqnarray}
However, if the data is high dimensional, we must resort to efficient numerical methods, similar as in SVMs.

The M-step is actually equivalent to minimizing the expectation of a re-weighted quadratic loss, as in Lemma~\ref{lemma:Corrupt-LR}. The proof is similar to that of Lemma 1 and the expectation of a Polya-Gamma distribution follows~\cite{Polson:arXiv12}.
\begin{lem}\label{lemma:Corrupt-LR}
Given $q(\lambdav)$, the M-step minimizes the re-weighted quadratic loss (with the $\ell_2$-norm regularizer)
\setlength\arraycolsep{1pt}\begin{eqnarray}
\Vert \wv \Vert_2^2 + \frac{c}{2} \sum_n \gamma_n^l \ep_{p}[ (\wv^\top \tilde{\xv}_n - y_n^l)^2 ],
\end{eqnarray}
where $y_n^l = \frac{c}{2 \gamma_n} y_n$ is the re-weighted label, $\gamma_n^l = \frac{\gamma_n}{c}$ and
\begin{eqnarray}\label{eq:ExpLambda2}
\gamma_n \triangleq \ep_{q}[\lambda_n] = \frac{c}{2\sqrt{\ep_p[\omega_n^2]}}\times\frac{e^{\sqrt{\ep_p[\omega_n^2]}}-1}{1+e^{\sqrt{\ep_p[\omega_n^2]}}}.
\end{eqnarray}
\end{lem}
It can be observed that if we fix $\gamma_n = \frac{c}{2}$, the IRLS algorithm reduces to minimizing the expected quadratic loss under the corrupting distribution. This is similar as in the case with SVMs, where if we set $\ell=0$ and fix $\gamma_n = \frac{1}{c}$, the IRLS algorithm for SVMs essentially minimizes the expected quadratic loss under the corrupting distribution.
Furthermore, by sharing a similar iterative structure, our IRLS algorithms shed light on the similarity and difference between the hinge loss and the logistic loss, as summarized in Table 1. Specifically, both losses can be minimized via iteratively minimizing the expectation of a re-weighted quadratic loss, while they differ in the update rules of the weights $\gamma_n$ and the labels $y_n$ at each iteration. 

\subsection{Dropout LR with Representation Learning}

Similar as in Section~\ref{sec:NLSVM}, we extend the logistic regression (LR) to learn latent representations under the dropout learning context.
Specifically, let $g(\xv; \alphav)\in \mathbb{R}^K$ denote the nonlinear transformation of the input features $\xv$, parameterized by $\alphav$. For the $D$-dimensional input $\xv$, let $K$ denote the transformed feature dimension. We again consider the logistic transformation, where each element $k$ is $g_k(\xv;\alphav) = \textrm{Sigmoid}( \alphav_k^\top \xv )$.
We then define our linear discriminant function\footnote{The offset is again ignored for simplicity.} as $$f(\xv; \wv, \alphav) = \wv^\top \gv(\xv; \alphav),$$
where $\wv \in \mathbb{R}^K$ is the classifier weights. We still let $\zeta_n = \wv^\top \gv(\xv; \alphav)$. Then, we have the same expected logistic loss as in Eq.~(\ref{eqn:EpLogistLoss}).

Using the same data augmentation technique, we can derive a variational upper bound of the expected logistic loss in Eq.~(\ref{eqn:EpLogistLoss}). 
We further need to deal with the nonlinear feature transformation, which renders the variational bound intractable. Here, we adopt the same strategy of using
first-order Taylor-expansion of $\gv(\cdot)$ as in Eq. (\ref{eqn:taylorExpension}) to get an approximation around the mean corrupted features $\muv_n = \ep_p[\xvtilde_n]$, 
and then compute the variational bound, which basically involves the variance of the corruption.
Then, an alternating minimization algorithm can be developed to iteratively update the following steps: 

{\bf For $q(\lambdav)$}: the solution has the same form as in (\ref{eq:lr-qlambda}):
\setlength\arraycolsep{1pt}\begin{eqnarray}
    q(\lambda_n) 
\mathfrak{}    &=&\mathcal{PG}\left( \lambda_n; c, \sqrt{\ep_{p}[\zeta_n^2]} \right),\nonumber
\end{eqnarray}
but with the new definition of $\zeta_n$. Under the Taylor-expansion, we have 
the second-order expectation as:
\begin{eqnarray}
    \ep_p[\zeta_n^2] =&&  \wv^\top(\ep_p[\gv^\prime(\tilde{\xv}_n) \gv^\prime(\tilde{\xv}_n)^\top]) \wv \nonumber \nonumber
\end{eqnarray}
where $\ep_p[\gv^\prime(\tilde{\xv}_n)] = \gv(\muv_n)$ and $\ep_p[\gv^\prime(\tilde{\xv}_n)\gv^\prime(\tilde{\xv}_n)^\top] = \gv(\muv_n)\gv(\muv_n)^\top + \triangledown_{\xvtilde} \gv(\muv_n)^\top V_p[\tilde{\xv}_{n}]\triangledown_{\xvtilde} \gv(\muv_n)$. 

{\bf For $\wv$}: this step is similar as in the linear case, but with subtle changes on the features. Specifically, by ignoring the irrelevant terms, we optimize the following objective: 
\begin{eqnarray}\label{eq:w-subproblem}
\mathcal{L}_{[\wv]} &=& \|\wv\|_2^2 
+ \sum_{n=1}^{N}\frac{1}{2}\gamma_n \ep_p[\zeta_n^2]-\frac{c}{2} y_n\ep_p[\zeta_n],
\end{eqnarray}
where $\gamma_n \triangleq \ep_q[\lambda_n]=\frac{c}{2\sqrt{\ep_p[\zeta_n^2]}}\times\frac{e^{\sqrt{\ep_p[\zeta_n^2]}}-1}
{1+e^{\sqrt{\ep_p[\zeta_n^2]}}}$. 
If $K$ is not large, we can solve this subproblem in a closed-form as in (\ref{eq:logist-analytical-solution}), with $\xvtilde_n$ replaced by $\gv^\prime(\tilde{\xv}_n)$.
If $K$ is large, we can resort to numerical methods (e.g., quasi-Newton methods) and use the derivative for $\wv$:
\begin{eqnarray}
\frac{\partial \mathcal{L}_{[\wv]}}{\partial w_k} &=& 2 w_k + c^2 \sum_n\gamma_n\!\Big((\tilde{y}_n - y_n^l) g_k(\muv_n) \nonumber \\
&& + \sum_d V_p[\tilde{x}_{nd}]\tilde{h}_{nd}\triangledown_{\tilde{\xv}} g_{dk}(\muv_n)\Big).\nonumber
\end{eqnarray}
where $\tilde{y}_n=\sum_k w_k g_k(\muv_n)$, $y_n^l=\frac{c}{2}\gamma_n^{-1}y_n$ and $\tilde{h}_{nd} = \sum_k w_k \triangledown_{\tilde{\xv}} g_{dk}(\muv_n)$. 

{\bf For $\alphav$}: this step involves optimizing the objective:
\begin{eqnarray}\label{eq:alpha-subproblem}
\mathcal{L}_{[\alphav]} &=& \|\alphav\|_2^2 + \sum_{n=1}^{N}\frac{1}{2}\gamma_n \ep_p[\zeta_n^2]-\frac{c}{2} y_n\ep_p[\zeta_n],
\end{eqnarray}
with a gradient descent method, where the gradient is
\begin{eqnarray}\label{eqn:NLSVM_alpha}
\frac{\partial \mathcal{L}_\alpha}{\partial \alpha_{dk}} &=& 2 \alpha_{dk}
+ c^2 \sum_n \gamma_n \Big((\tilde{y}_n - y_n^l)\mu_{nd} \nonumber \\
&& + V_p[\tilde{x}_{nd}]\tilde{h}_{nd}\rho_{kd}\Big)w_k \eta_k,\nonumber
\end{eqnarray}
and $\rho_{kd}$ and $\eta_k$ are the same as in Section~\ref{sec:NLSVM}. 

In summary, the nonlinear logistic regression can be learned via coordinate descent by iteratively updating $\gammav$, solving for model parameters $\wv$ by minimizing $\mathcal{L}_{[\wv]}$ in (\ref{eq:w-subproblem}), and solving for the transformation weights $\alphav$ by minimizing $\mathcal{L}_{[\alphav]}$ in (\ref{eq:alpha-subproblem}).

\section{Experiments}\label{sec:experiment}

We now present empirical results on classification, regression and the challenging ``nightmare at test time" scenario~\cite{Globerson:icml06} to demonstrate the effectiveness of the dropout training algorithm for SVMs, denoted by (linear) Dropout-SVM and its nonlinear version Dropout-LatentSVM, and the new IRLS algorithms for the dropout training of the logistic loss, denoted by (linear) Dropout-LR and its nonlinear version Dropout-LatentLR.

\subsection{Datasets and Settings}

We evaluate our proposed models for classification and regression on 9 datasets, including 1) {\it Amazon review}~\cite{AmazonReviewData}: four types of product text review datasets including books, kitchen, dvd and electronics. The binary classification task is to distinguish whether a review content is positive or negative; 2) {\it Dmoz}: a large collection of webpages organized in a tree hierarchy with 16 categories; 3) {\it Reuters}: a dataset with the documents appeared on the Reuters newswire in 1987 with 65 categories; 4) {\it CIFAR}:\footnote{http://www.cs.toronto.edu/$\sim$kriz/cifar.html} the subset of the 80 million tiny images \cite{Torralba:PAMI08}. It consists of 10 classes of $32\times 32$ tiny images. We follow the experimental setup of the previous work~\cite{Cifar-10Data,CorruptICML2013}; 5) {\it MNIST}: a dataset that consists of 60,000 training and 10,000 testing handwritten digital images from 10 categories (i.e., $0, \cdots, 9$). The images are represented by $28\times 28$ pixels which results in the feature dimension of 784; and 6) {\it Hotelreview}~\cite{CTRF:2010}: a dataset that consists of 5,000 hotel reviews randomly collected from TripAdvisor. Each document is associated with a global rating score, ranging from 1 to 5. We normalize the rating scores as in~\cite{CTRF:2010} for regression. Table~\ref{table:dataset} summarizes the statistics of these datasets.

\begin{table}[t]\vspace{-.2cm}
\caption{A summary of the 9 datasets.}\label{table:dataset}\vspace{-.3cm}
\begin{center}
 \scalebox{1}
 { \setlength{\tabcolsep}{1.8pt}
       \begin{tabular}{|c|c|c|c|c|}
       \hline
        \hline
        Dataset           &  Train Size  & Test Size & Feature Dim & Categories \\
        \hline
        Amazon-books        & 2,000 & 4,465 & 20,000 & 2 \\
        Amazon-kitchen        & 2,000 & 5,945 & 20,000 & 2 \\
        Amazon-dvd        & 2,000 & 3,586 & 20,000 & 2 \\
        Amazon-electronics  & 2,000 & 5,681 & 20,000 & 2 \\
        \hline
        Dmoz     & 7184 & 1796 & 16,498 & 16 \\
        Reuters        & 5,946 & 2,347 & 18,933 & 65 \\
        CIFAR-10     & 50,000 & 10,000 & 8,192 & 10 \\
        MNIST     & 60,000 & 10,000 & 784 & 10 \\
        \hline
        Hotelreview     & 2,500 & 2,500 & 12,000 & Regression \\
        \hline
        \end{tabular}
}
\end{center}\vspace{-.5cm}
\end{table}

We consider the unbiased dropout (or blankout) noise model\footnote{Other noise models (e.g., Poisson) were shown to perform worse than the dropout model~\cite{CorruptICML2013}. We have similar observations for Dropout-SVM and the new IRLS algorithm for logistic regression.}, that is,
$p(\tilde{\xv}=0)= q$ and $p(\tilde{\xv}=\frac{1}{1-q}\xv)= 1-q$, where $q \in [0,1)$ is a pre-specified corruption level. Then, the variance for each dimension $d$ is $V_p[\tilde{x}_d] = \frac{q}{1-q} x_d^2$.

\begin{figure}[h]\vspace{-.1cm}
\begin{center}
\includegraphics[width=.65\linewidth]{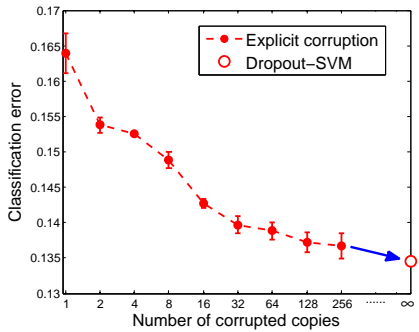}\vspace{-0.4cm}
\end{center}
\caption{Comparison between Dropout-SVM and the explicit corruption for SVM on the Amazon-books dataset.}
\label{fig:expSVM}\vspace{-0.4cm}
\end{figure}

\begin{figure*}
\centering
\subfigure[books]{\includegraphics[height=1.55in, width=1.6in]{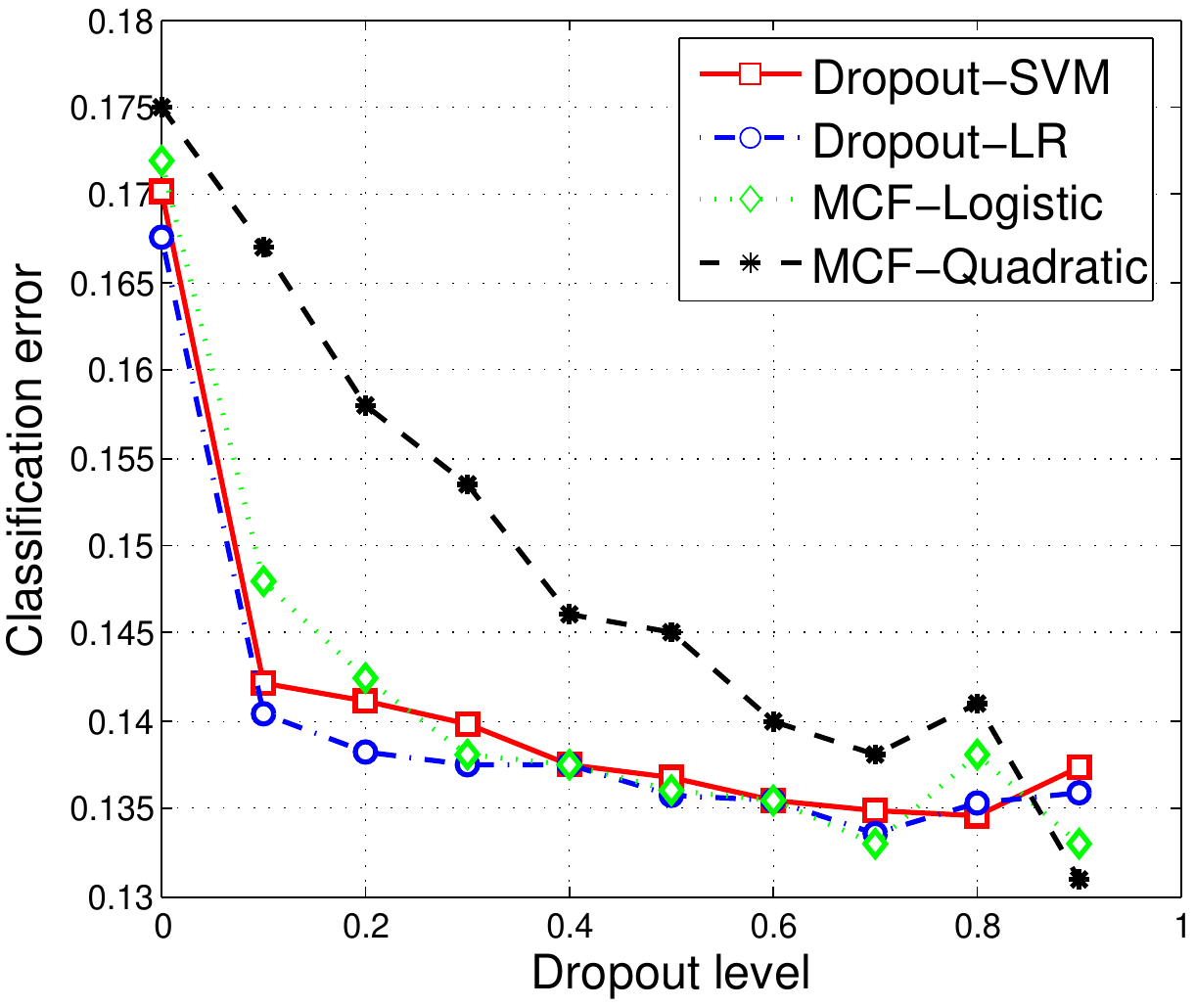}}\label{fig:amazon_books}
\subfigure[kitchen]{\includegraphics[height=1.55in, width=1.6in]{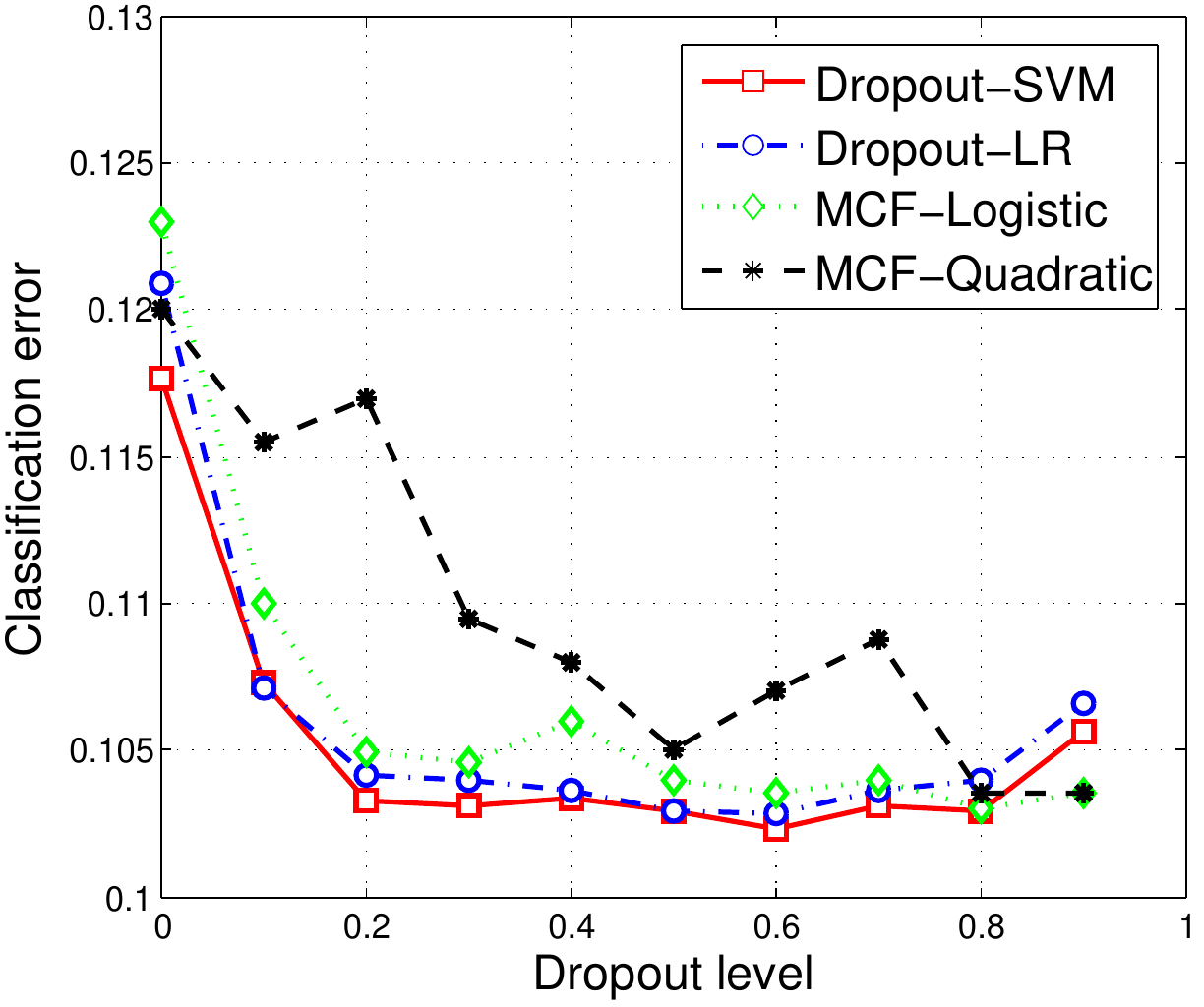}}\label{fig:amazon_kitchen}
\subfigure[dvd]{\includegraphics[height=1.55in, width=1.6in]{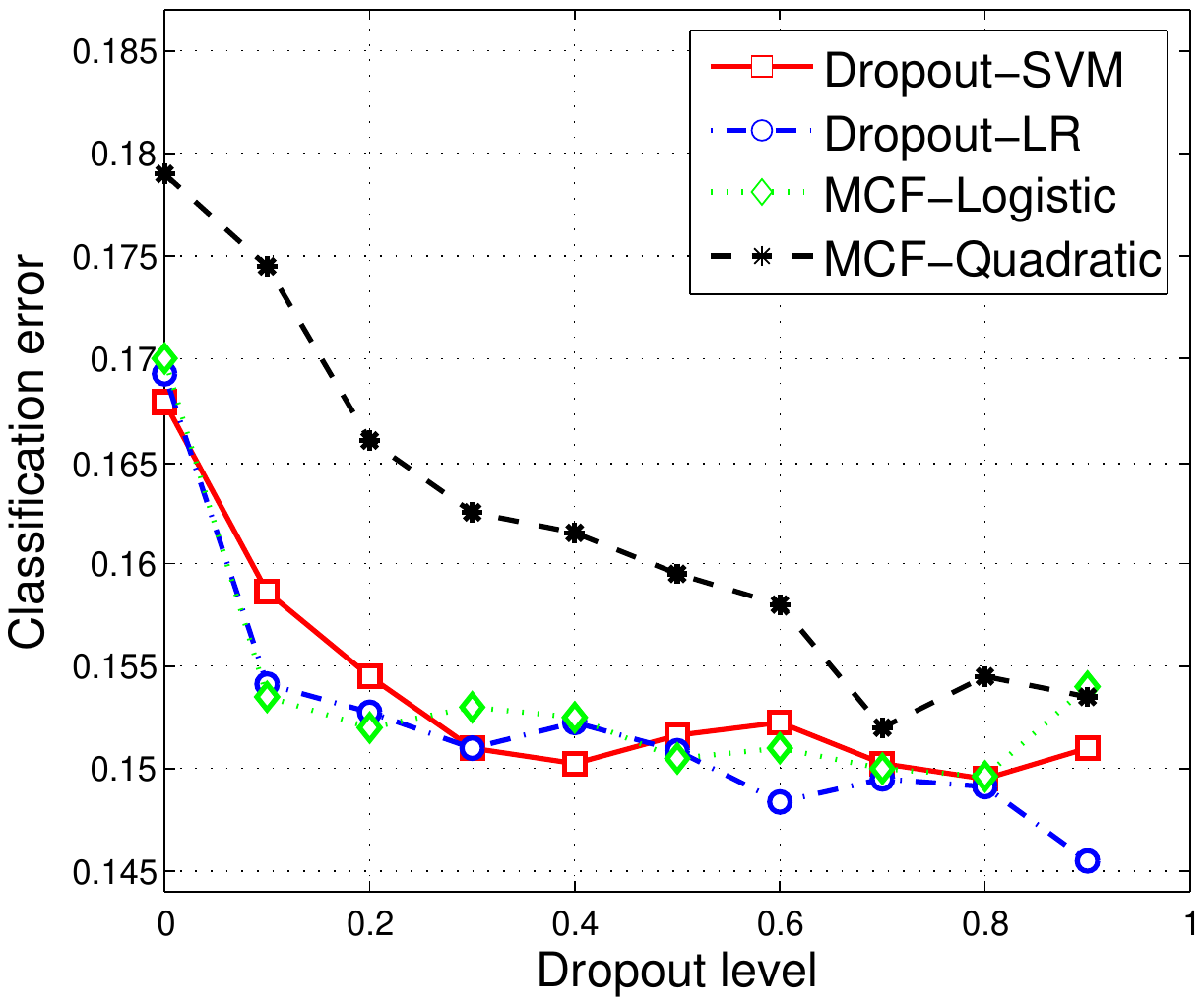}}\label{fig:amazon_dvd}
\subfigure[electronics]{\includegraphics[height=1.55in, width=1.6in]{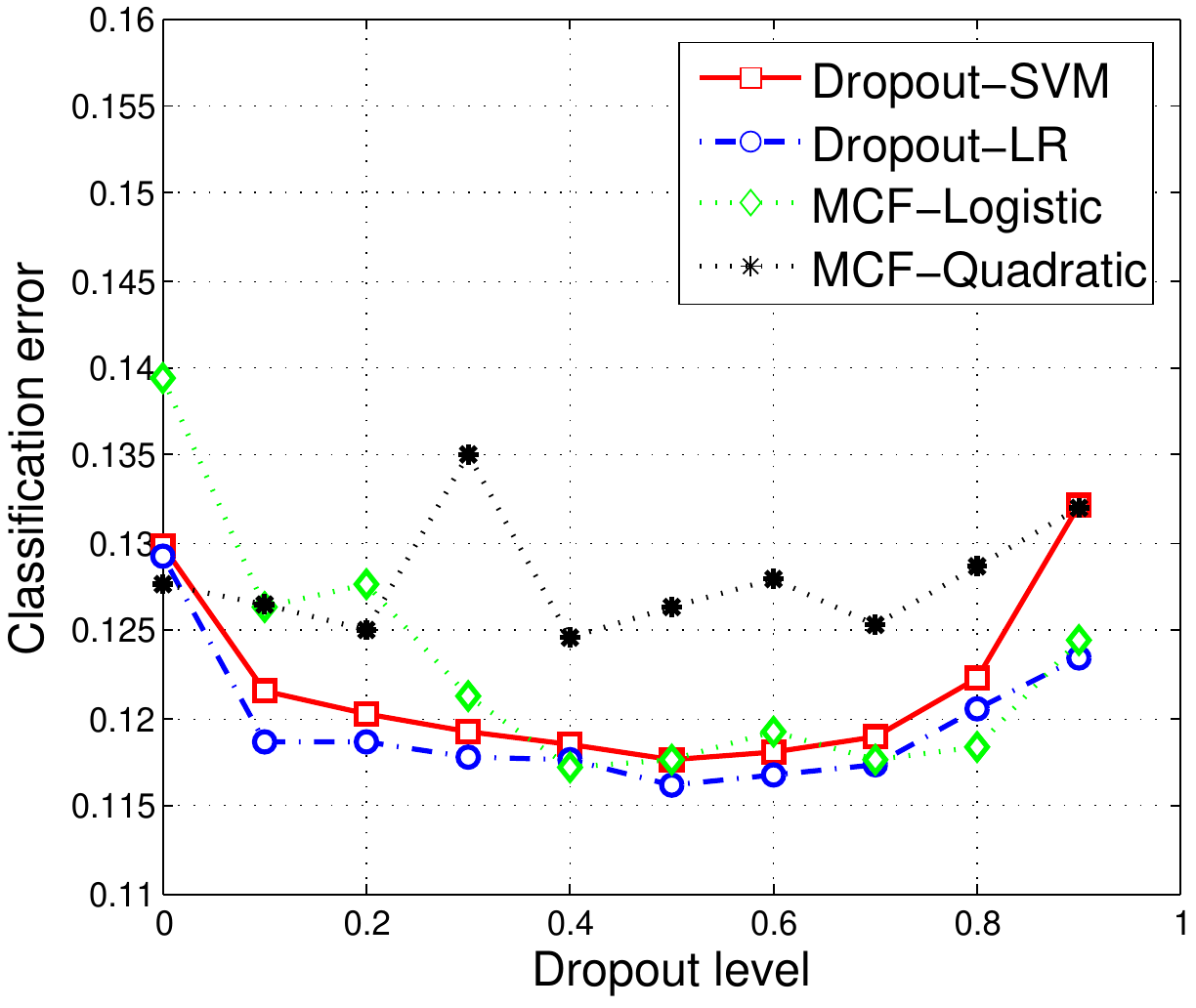}}\label{fig:amazon_electronics}\vspace{-.2cm}
\caption{Classification errors on the Amazon datasets. Best viewed in color. }
\label{fig:amazon}\vspace{-.3cm}
\end{figure*}

\subsection{Linear Dropout Classifiers}

We first compare the marginalized corruption of Dropout-SVM with the explicit corruption strategy for SVM, and then evaluate our linear dropout classifiers on both binary and multi-class classification to show the effectiveness of dropout training on linear SVMs and logistic regression.

\subsubsection{Dropout-SVM vs. Explicit corruption}


Fig.~\ref{fig:expSVM} shows the classification errors on the Amazon-books dataset when a SVM classifier is trained using the explicit corruption strategy as in~Eq. (\ref{eq:ExplicitCorruption}). We change the number of corrupted copies (i.e., $M$) from $1$ to $256$. Following the previous setups~\cite{CorruptICML2013}, for each value of $M$ we choose the dropout model with $q$ selected by cross-validation. The hyper-parameter of the SVM classifier is also chosen via cross-validation on the training data. We can observe a clear trend that the error decreases when the training set contains more corrupted versions of the original training data, i.e., $M$ gets larger in Eq.~(\ref{eq:ExplicitCorruption}). It also shows that the best performance is obtained when $M$ approaches infinity, which is equivalent to our Dropout-SVM.


\subsubsection{Binary classification}

We evaluate Dropout-SVM and Dropout-LR on binary classification tasks. We use the four Amazon review datasets as detailed in Table~\ref{table:dataset}.
The task is to distinguish whether a review content is positive or negative.


We compare our methods with those presented in~\cite{CorruptICML2013} that minimize the quadratic loss with marginalized corrupted features (MCF), denoted by MCF-Quadratic, and that minimize the expected logistic loss, denoted by MCF-Logistic. MCF-Logistic was shown to be the state-of-the-art method for dropout training on these datasets, outperforming a wide range of competitors, including the dropout training of the exponential loss and the various loss functions with a Poisson noise model. As we have discussed, both Dropout-SVM and Dropout-LR iteratively minimize the expectation of a re-weighted quadratic loss, with the re-weights updated in closed-form. We include MCF-Quadratic as a baseline to demonstrate the effectiveness of our methods on adaptively tuning the re-weights to get improved results. We implement both Dropout-SVM and Dropout-LR using C++, and solve the re-weighted least square problems using L-BFGS methods~\cite{Liu:89}, which are very efficient by exploring the sparsity of bag-of-words features when computing gradients\footnote{We don't compare time with MCF methods, whose implementation (http://homepage.tudelft.nl/19j49/mcf/Marginalized\_Corrupted\_Features.html) are in Matlab, slower than ours.}.


Fig.~\ref{fig:amazon} shows classification errors, where we cite the results of MCF-Logistic and MCF-Quadratic from~\cite{CorruptICML2013}. We can see that on all datasets, Dropout-SVM and Dropout-LR generally outperform MCF-Quadratic except when the dropout level is larger than 0.9, suggesting that adaptively updating the re-weights can improve the performance. In the meanwhile, the proposed two models give comparable results with (a bit better than on the kitchen dataset) the state-of-art MCF-Logistic which means that dropout training on SVMs is an effective strategy for binary classification. Finally, by noting that Dropout-SVM reduces to the standard SVM when the corruption level $q$ is zero, we can see that dropout training can significantly boost the classification performance for the simple linear SVMs.



\subsubsection{Multi-class classification}

We also evaluate on multiclass document/image classification tasks, using DMOZ, Reuters and CIFAR-10 datasets.
There are various approaches to applying binary Dropout-SVM and Dropout-LR to multiclass classification, including ``one-vs-all" and ``one-vs-one" strategies. Here we choose ``one-vs-all", which has shown effectiveness in many applications~\cite{InDefense:JMLR04}. The hyper-parameters are selected via cross-validation on the training set. 

\begin{figure}\vspace{-.2cm}
\centering
\subfigure[DMOZ dataset]{\includegraphics[height=1.55in, width=1.6in]{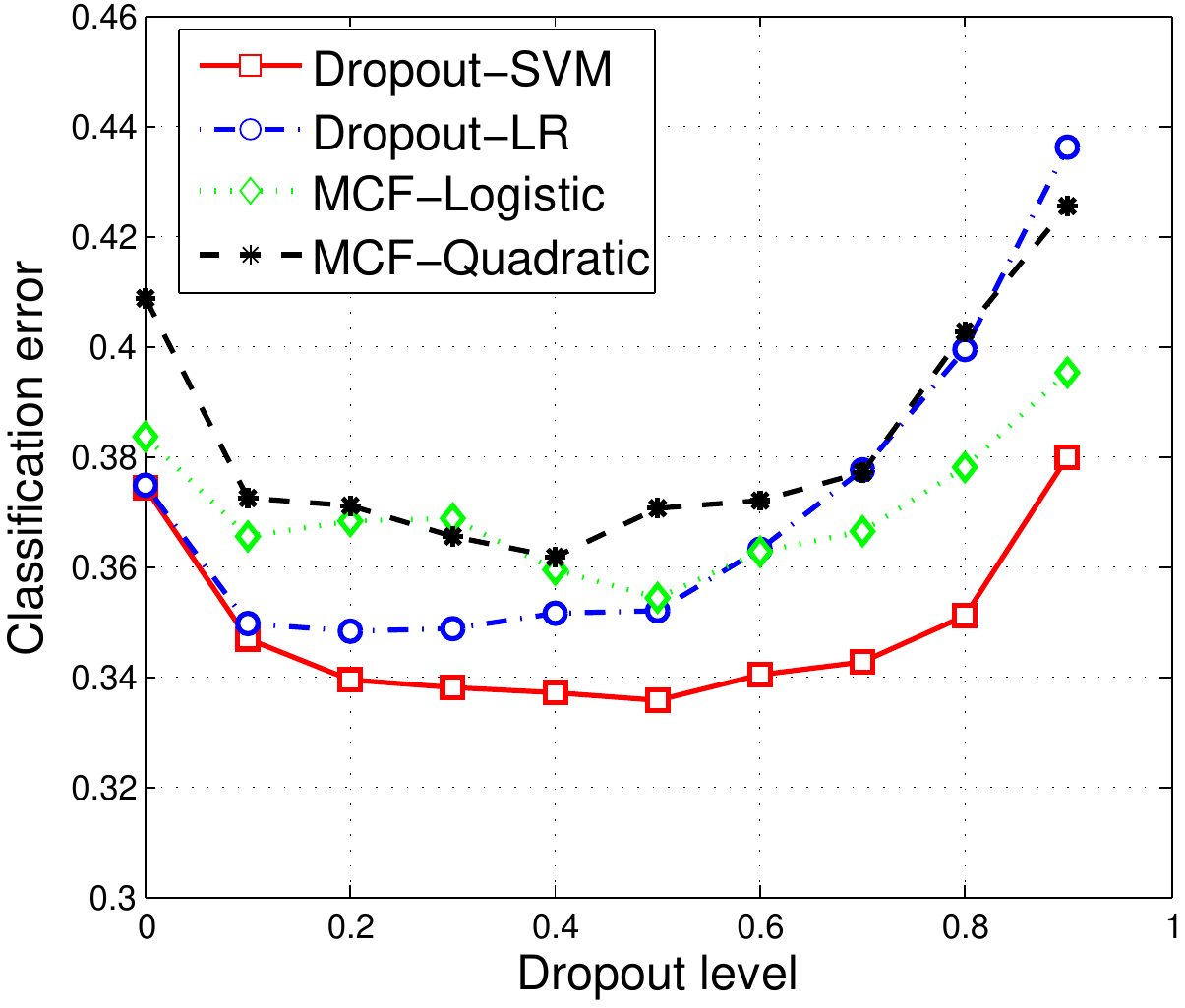}\label{fig:DMOZ}}
\subfigure[Reuters dataset]{\includegraphics[height=1.55in, width=1.6in]{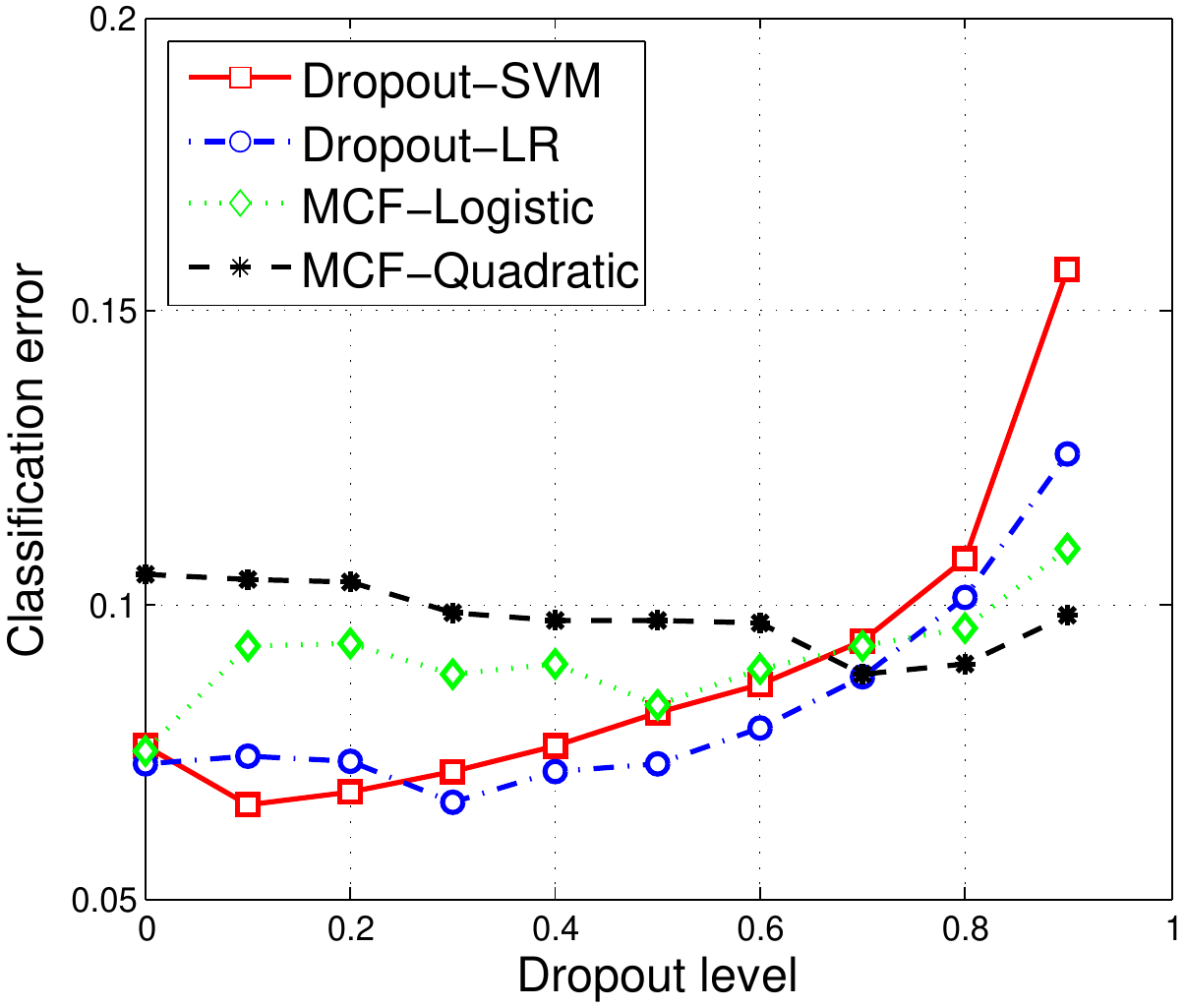}}\vspace{-.2cm}
\caption{Classification errors on Dmoz and Reuters datasets.}\label{fig:DmozAndReuters}
\label{fig:DmozReuters}\vspace{-.3cm}
\end{figure}

{\bf Document classification}: Fig.~\ref{fig:DmozAndReuters} shows the classification errors on the DMOZ and Reuters dataset. It can be observed that all methods can successfully boost the performance with dropout training; besides, Dropout-SVM performs comparably with Dropout-LR (or better on DMOZ dataset) for different dropout levels, and significantly outperforms MCF-Quadratic, which demonstrates the effect of updating reweights in our IRLS algorithm. Moreover, Dropout-SVM performs slightly better than the state-of-the-art method (i.e., MCF-Logistic) under the dropout training setting. This is consistent with the binary classification observations.


{\bf Image classification}: Table~\ref{table:cifar10} presents the results on CIFAR-10 image dataset, where the results of quadratic loss and logistic loss under the MCF learning setting\footnote{The exponential loss was shown to be worse; thus omitted.} are cited from \cite{CorruptICML2013}. 
We can see that all the methods (except for the quadratic loss) can significantly boost the performance by adopting dropout training. Meanwhile, both Dropout-SVM and Dropout-LR are competitive, in fact achieving comparable performance as the state-of-the-art method (i.e., MCF-Logistic) under the dropout training setting. Both Dropout-SVM and Dropout-LR outperform MCF-Quadratic which demonstrate the effect of updating reweights in our IRLS algorithm. 

\begin{table}[t]\vspace{-.1cm}
\caption{Classification errors on the CIFAR-10 dataset.}\label{table:cifar10}\vspace{-.3cm}
\begin{center}
 \scalebox{1.1}
 { \setlength{\tabcolsep}{1.8pt}
       \begin{tabular}{|c|c|c|c|}
       \hline
        \hline
        Model           &  No Corrupt  & Dropout & Dropout\\
        &&$q=0.2$&$q=0.3$\\
        \hline
        Dropout-SVM        & 0.322 & {\bf 0.291} & 0.293 \\
        Dropout-LR     & 0.312 & 0.291 & {\bf 0.290}\\
        MCF-Logist      & 0.325 & 0.296 & 0.294\\
        MCF-Quadratic     & 0.326 & 0.322 & 0.321\\
        \hline
        \end{tabular}
}
\end{center}\vspace{-.5cm}
\end{table}

\subsubsection{Regression}

We evaluate the dropout support vector regression (SVR) model on predicting rating scores for hotel review dataset. We compare with the MCF-quadratic model~\cite{CorruptICML2013}, which refers to the standard least square with marginalized dropout training. We use predictive R$^2$ as the measurement~\cite{Blei:07}, which is defined as $ R^2 \triangleq 1-\frac{\sum_d(y_d - \hat{y}_d)^2}{\sum_d(y_d-\bar{y})^2}$, where $y_d$ is the ground-truth response, $\hat{y}_d$ is the predicted value, and $\bar{y}$ is the mean of all the responses.
\begin{figure}
\centering
\includegraphics[height=1.6in, width=.65\linewidth]{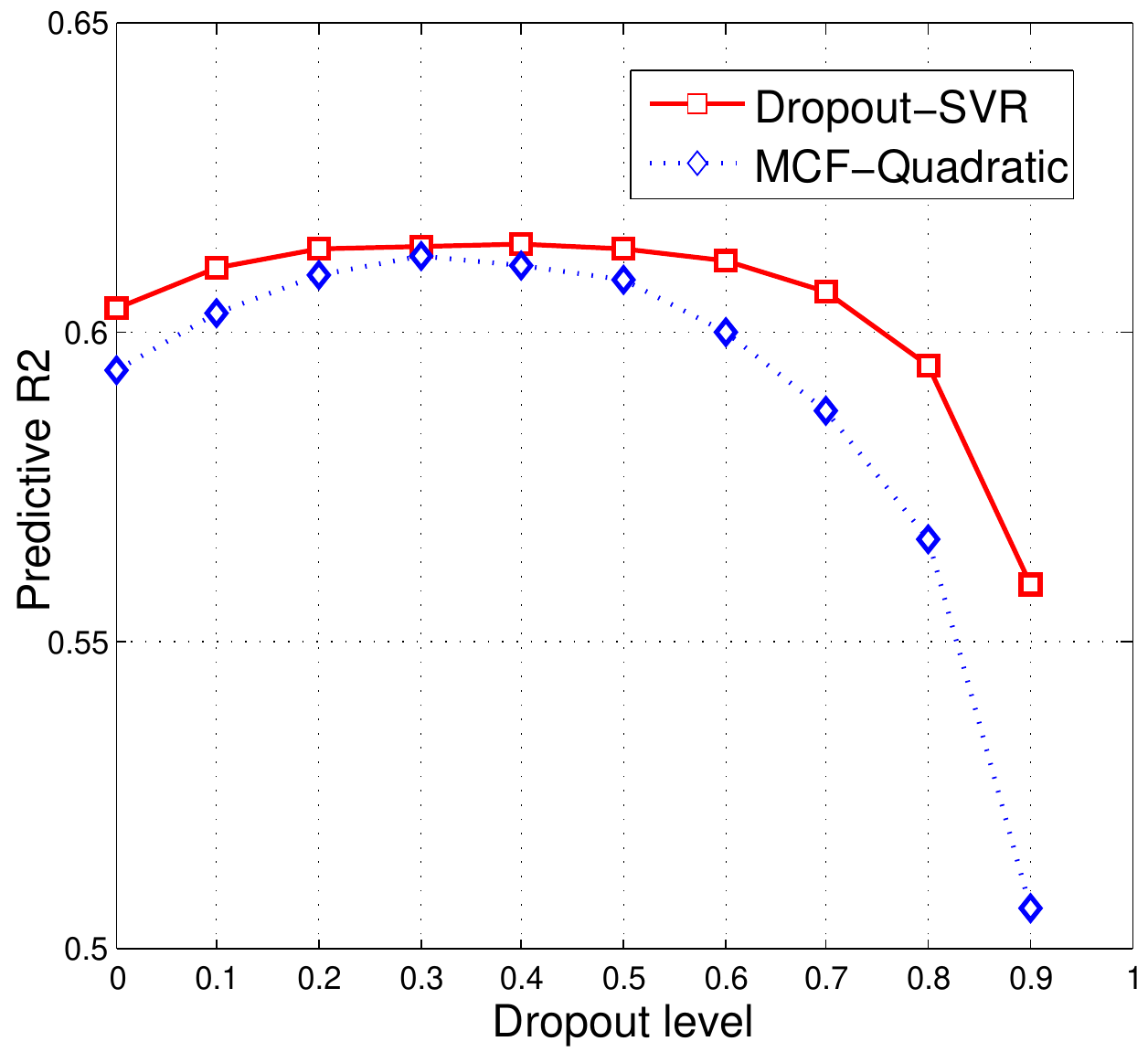}\vspace{-.2cm}
\caption{Prediction R2 on the Hotelreview dataset.}\label{fig:SVR_HotelReview}
\end{figure}

\begin{figure*}\vspace{-.2cm}
\centering
\subfigure[books]{\includegraphics[height=1.5in, width=1.6in]{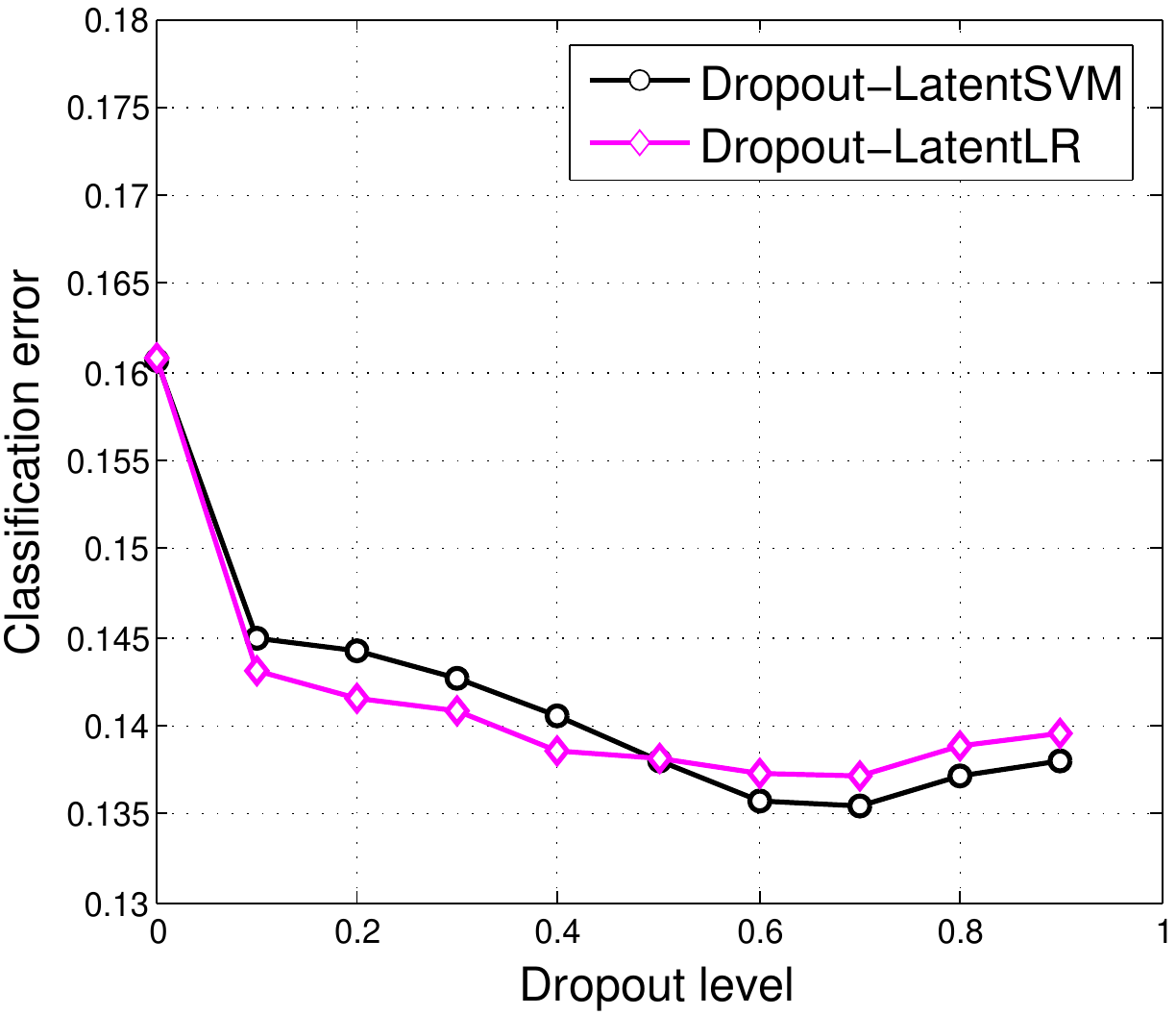}\label{fig:NL_books}}
\subfigure[kitchen]{\includegraphics[height=1.5in, width=1.6in]{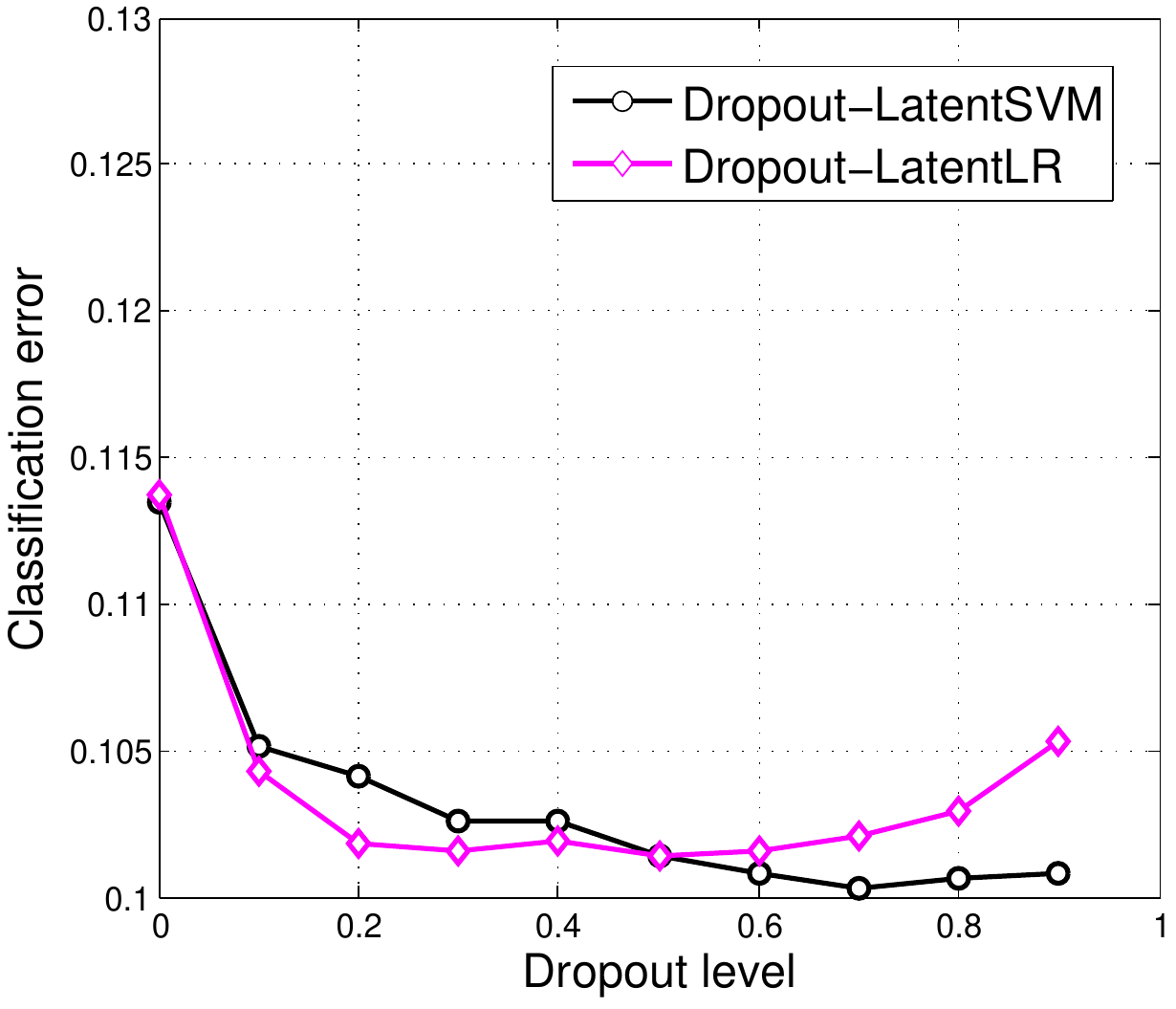}\label{fig:NL_kitchen}}
\subfigure[dvd]{\includegraphics[height=1.5in, width=1.6in]{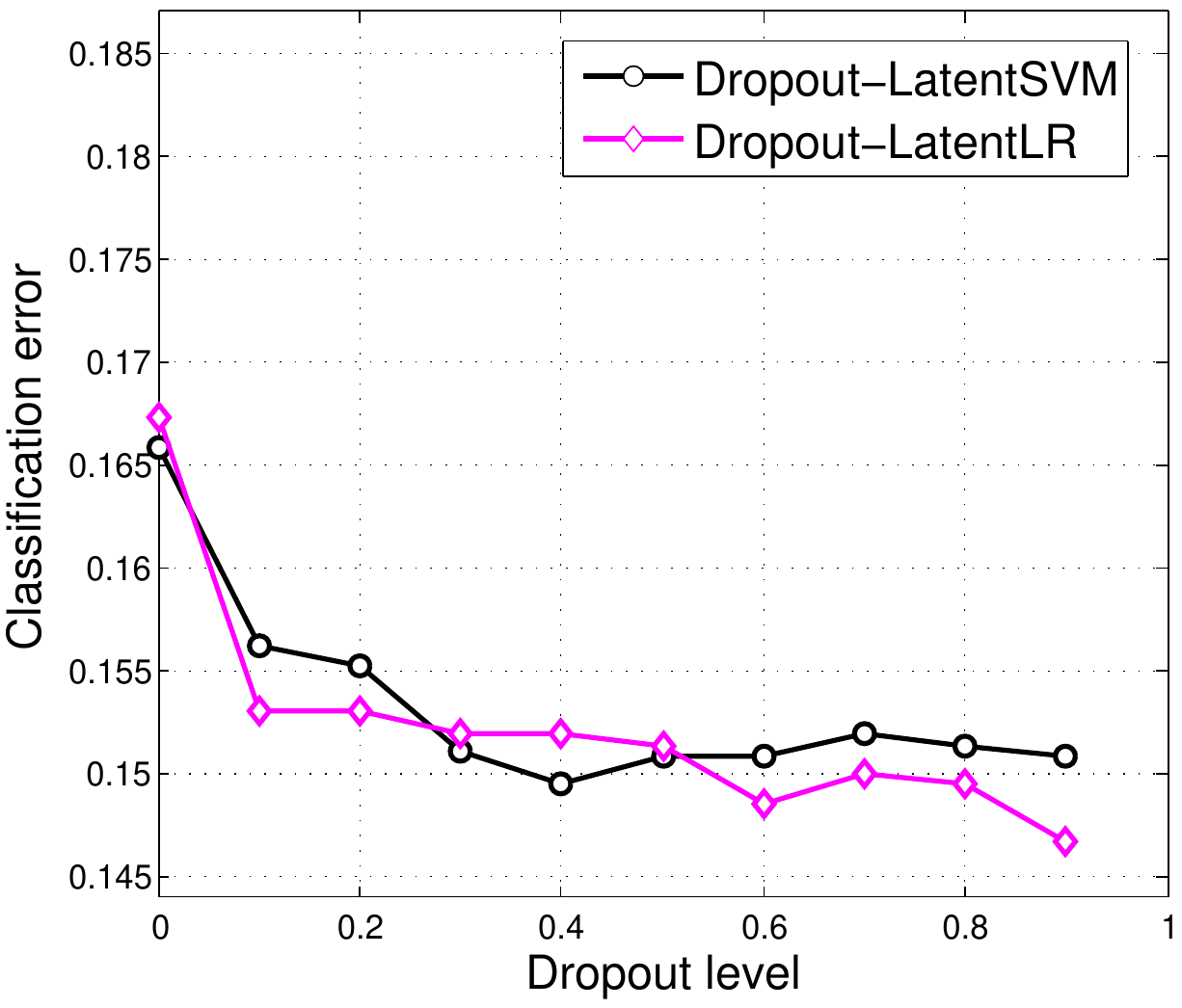}\label{fig:NL_dvd}}
\subfigure[electronics]{\includegraphics[height=1.5in, width=1.6in]{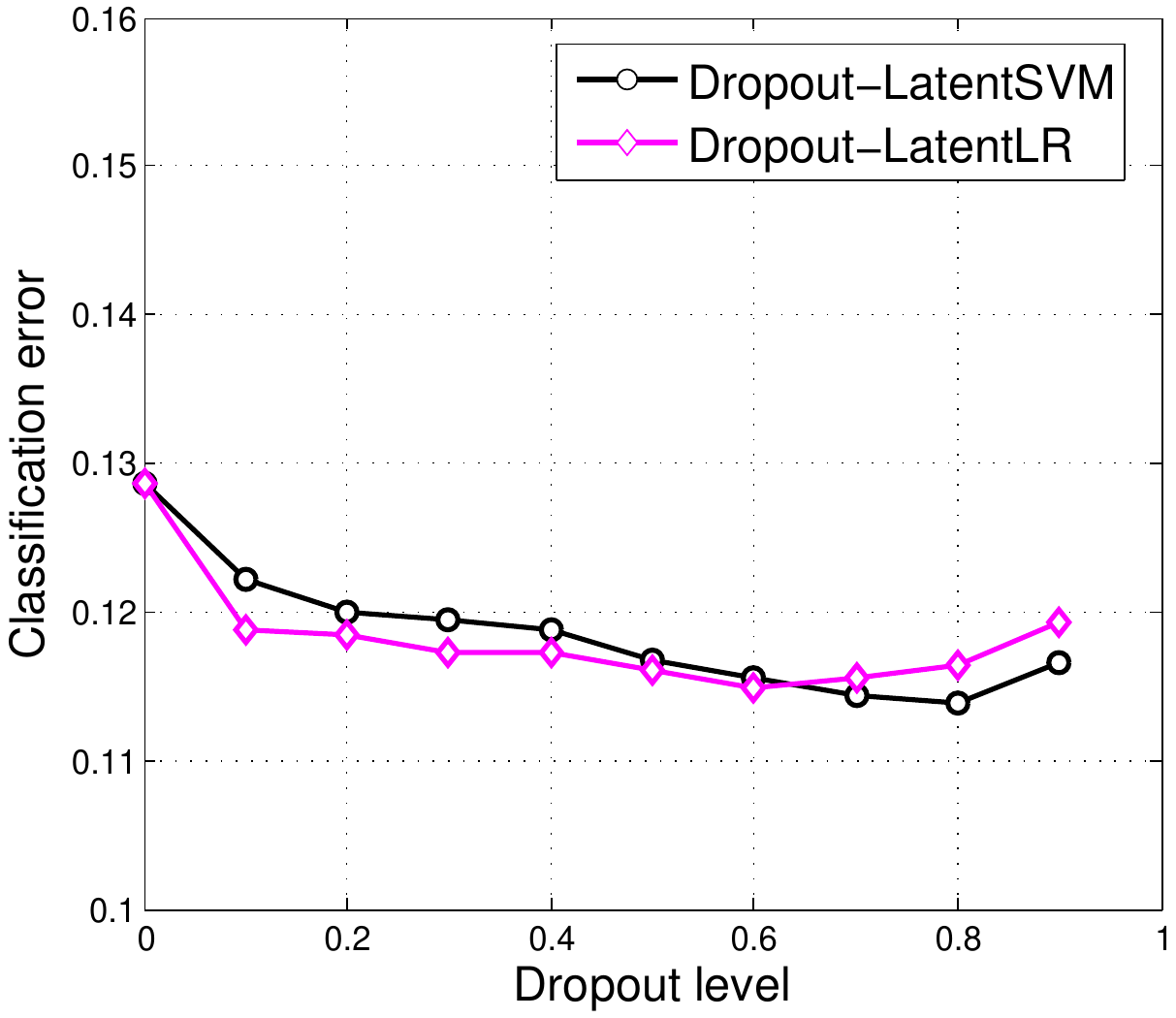}\label{fig:NL_electronics}}\vspace{-.2cm}
\caption{Classification errors of Dropout-LatentSVM and Dropout-LatentLR on the Amazon datasets. }
\label{fig:NLCorrupt}\vspace{-.3cm}
\end{figure*}

Fig.~\ref{fig:SVR_HotelReview} shows the predictive R$^2$ score. We can see: 1) Dropout-SVR outperforms MCF-Quadratic for all the dropout levels, which implies the discriminative power of our Dropout-SVR using the $\epsilon$-insensitive loss; 2) noting that Dropout-SVR reduces to the standard SVR when the corruption level $q$ is zero, Dropout-SVR can successfully improve the performance except when corruption level $q$ is larger than 0.8. This means that our dropout training strategy is also effective for regression tasks. The best regression performance is obtained when $0.1<q<0.5$.

\subsection{Classifiers with Representation Learning}

We evaluate the Dropout-LatentSVM and Dropout-LatentLR on both image and text classification tasks.

For image classification, Table~\ref{table:cifar10-NL} shows the errors of different nonlinear models on CIFAR-10 dataset. We can see that the prediction performance is significantly improved by the nonlinear Dropout-SVM and Dropout-LR, especially when using dropout training. The best performance is obtained when latent dimension is 40. These  results demonstrate that our Dropout training strategy is very effective for nonlinear classifiers.

For text classification, Fig.~\ref{fig:NLCorrupt} shows the errors of Dropout-LatentSVM and Dropout-LatentLR on Amazon review datasets. We have following observations: 1) both methods perform comparably on the four datasets, which is not surprising due to the very similar IRLS algorithms; 
2) dropout training can consistently boost the classification performance for both Dropout-LatentSVM and Dropout-LatentLR, compared with the standard nonlinear classifiers when the dropout level $q$ equals to zero; 3) the nonlinear Dropout classifiers do not obtain significant improvements compared with the linear classifiers on the document classification task, probably because the words are already high-level representations or the simple fully connected network is not suitable for text documents.

\begin{table}[t]\vspace{-.2cm}
\caption{Classification errors on the CIFAR-10 dataset.}\label{table:cifar10-NL}\vspace{-.3cm}
\begin{center}
 \scalebox{1}
 { \setlength{\tabcolsep}{1.8pt}
       \begin{tabular}{|c|c|c|c|}
       \hline
        \hline
        Model           &  No Corrupt  & Dropout & Dropout\\
        &&$q=0.2$&$q=0.3$\\
        \hline
        Dropout-LatentSVM (K = 10)& 0.292 & 0.287 & 0.285\\
        Dropout-LatentSVM (K = 20) & 0.291 & 0.275 & 0.271\\
        Dropout-LatentSVM (K = 30) & 0.289 &{\bf 0.269} & {\bf 0.266}\\
        Dropout-LatentSVM (K = 40) & 0.288 &{\bf 0.268}& {\bf 0.265}\\
        \hline
        Dropout-LatentLR (K = 10) & 0.295 &0.280 &0.277 \\
        Dropout-LatentLR (K = 20) & 0.290 &0.271 &0.268 \\
        Dropout-LatentLR (K = 30) & 0.285 &{\bf 0.264} &{\bf 0.264} \\
        Dropout-LatentLR (K = 40) & 0.284 &{\bf 0.262} &{\bf 0.261} \\
        \hline
        \end{tabular}
}
\end{center}\vspace{-.5cm}
\end{table}

\subsection{Nightmare at test time}

Finally, we evaluate our methods under the ``nightmare at test time"~\cite{Globerson:icml06} supervised learning scenario, where some input features that were present when building the classifiers may ``die" or be deleted at testing time. In such a scenario, it is crucial to design algorithms that do not assign too much weight to any single feature during testing, no matter how informative it may seem at training. Previous work has conducted the worst-case analysis as well as the learning with marginalized corrupted features. We take this scenario to test the robustness of our dropout training algorithms for linear Dropout-SVM, Dropout-LR as well as the nonlinear Dropout-LatentSVM and Dropout-LatentLR.

We follow the setup of \cite{CorruptICML2013} and choose 
the MNIST dataset. 
We train the models on the full training set, and evaluate the performance on different versions of test set in which a certain level of the features are randomly dropped out, i.e., set to zero. We compare the performance of our dropout learning algorithms with the state-of-art MCF-predictors that use the logistic loss and quadratic loss. These two models also show the state-of-art performance on the same task to the best of our knowledge. We also compare with FDROP~\cite{Globerson:icml06}, which is a state-of-the-art algorithm for the ``nightmare at test time" setting that minimizes the hinge loss under an adversarial worst-case analysis. During training, we choose the best models over different dropout levels via cross-validation. For both Dropout-SVM and Dropout-LR, we adopt the ``one-vs-all" strategy as above for the multiclass classification task.

\begin{figure}\vspace{-.2cm}
\centering
\subfigure[Linear Models]{\includegraphics[height=1.6in, width=1.6in]{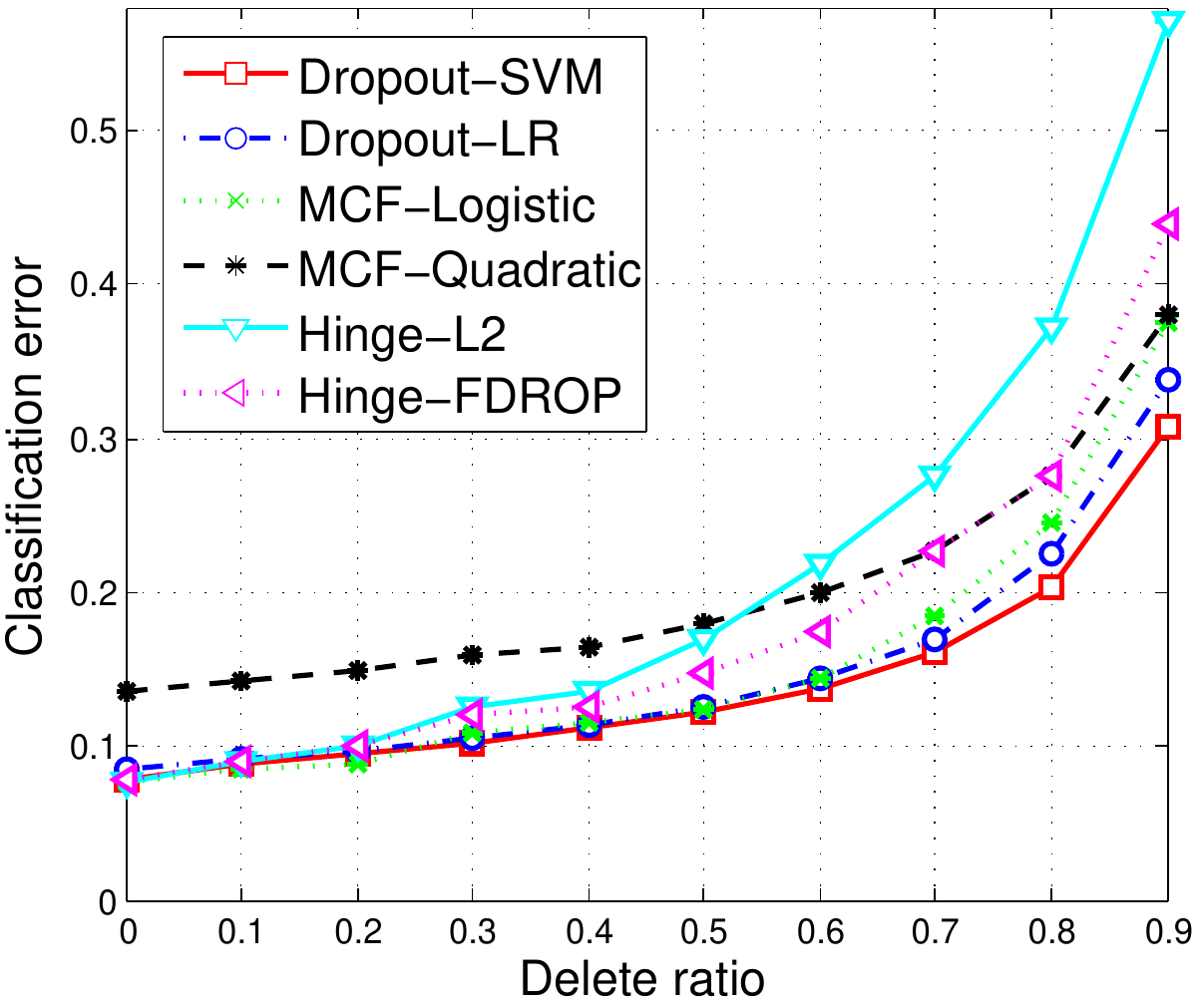}\label{CorruptSVM}}
\subfigure[Non-linear Models]{\includegraphics[height=1.6in, width=1.6in]{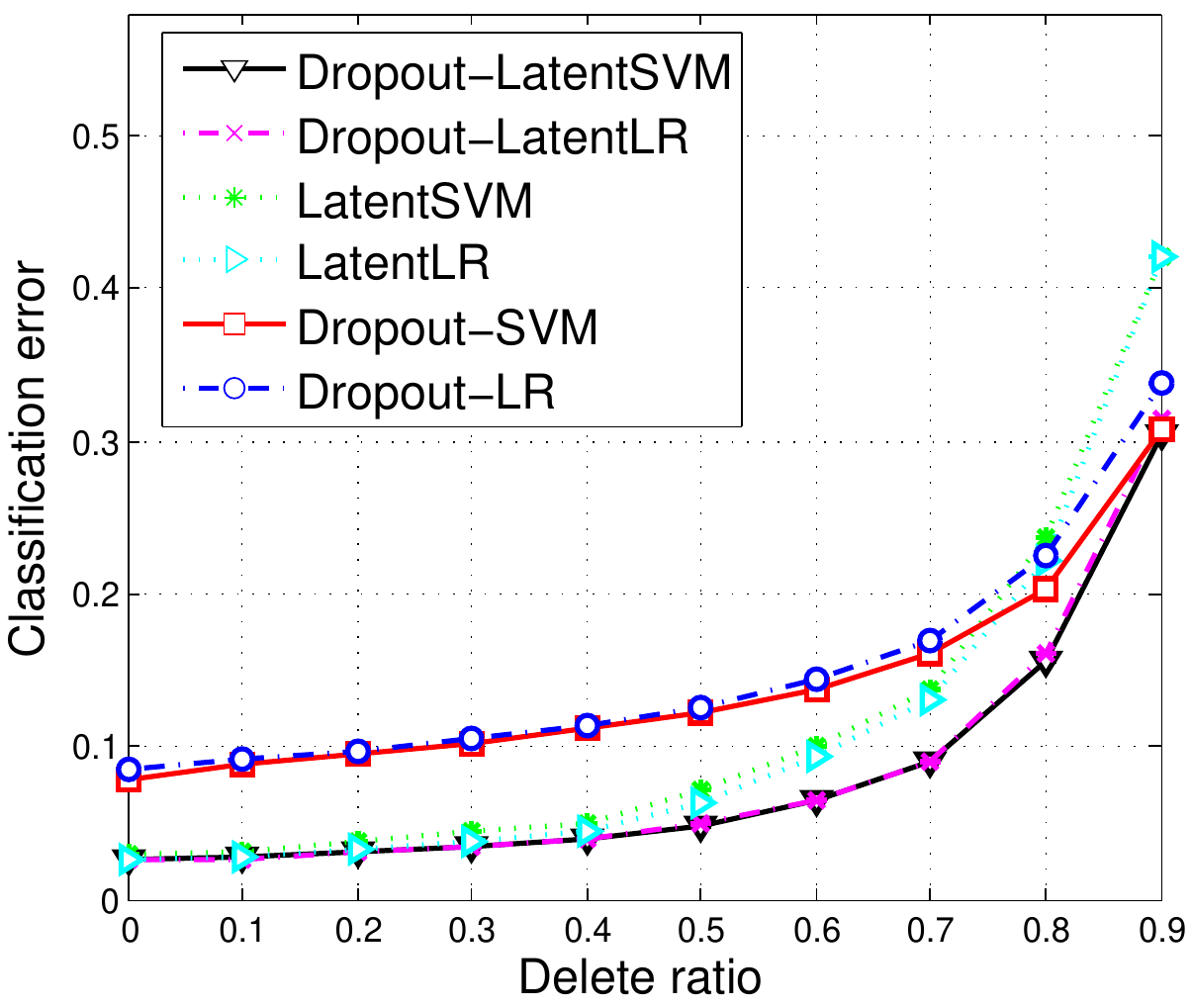}\label{NLCorruptSVM}}\vspace{-.2cm}
\caption{Classification errors of nightmare at test time on MNIST dataset.}
\label{fig:mnist}\vspace{-.4cm}
\end{figure}

Fig.~\ref{CorruptSVM} shows the classification errors of linear Dropout classifiers compared with other state-of-the-art methods as a function of the random deletion percentage of features at the testing time. Following previous settings, for each deletion percentage, we use a small validation set with the same deletion level to determine the regularization parameters and the dropout level $q$ on the whole training data. From the results, we can see that the proposed Dropout-SVM is consistently more robust than all the other competitors, including the two methods to minimize the expected logistic-loss, especially when the feature deletion percentage is high (e.g., $> 50\%$). Comparing with the standard SVM (i.e., the method Hinge-L2) and the worst-case analysis of hinge loss (i.e., Hinge-FDROP), Dropout-SVM consistently boosts the performance when the deletion ratio is greater than $10\%$. As expected, Dropout-SVM also significantly outperforms the MCF method with a quadratic loss (i.e., MCF-Quadratic), which is a special case of Dropout-SVM as shown in our theory. Finally, we also note that our iterative algorithm for the logistic-loss works slightly better than the previous algorithm (i.e., MCF-Logistic) when the deletion ratio is larger than $50\%$.

\begin{figure*}
 \centering
 \scalebox{.7}
 {
 \setlength{\tabcolsep}{2.5pt}
 \begin{tabular}{c| c c c c c c c c}
 \hline \hline
{\multirow{8}{*}{{\color{red} \bf positive}}} &
{\multirow{4}{*}{$k = 2$, $w_k = 17.734$}} &
{\multirow{4}{*}{\includegraphics[height=.18\columnwidth]{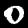}}} &
{\multirow{4}{*}{\includegraphics[height=.18\columnwidth]{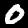}}} &
{\multirow{4}{*}{\includegraphics[height=.18\columnwidth]{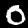}}} &
{\multirow{4}{*}{\includegraphics[height=.18\columnwidth]{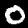}}} &
{\multirow{4}{*}{\includegraphics[height=.18\columnwidth]{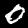}}} &
{\multirow{4}{*}{\includegraphics[height=.18\columnwidth]{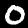}}} &
{\multirow{4}{*}{\includegraphics[width=.5\columnwidth,height=.19\columnwidth]{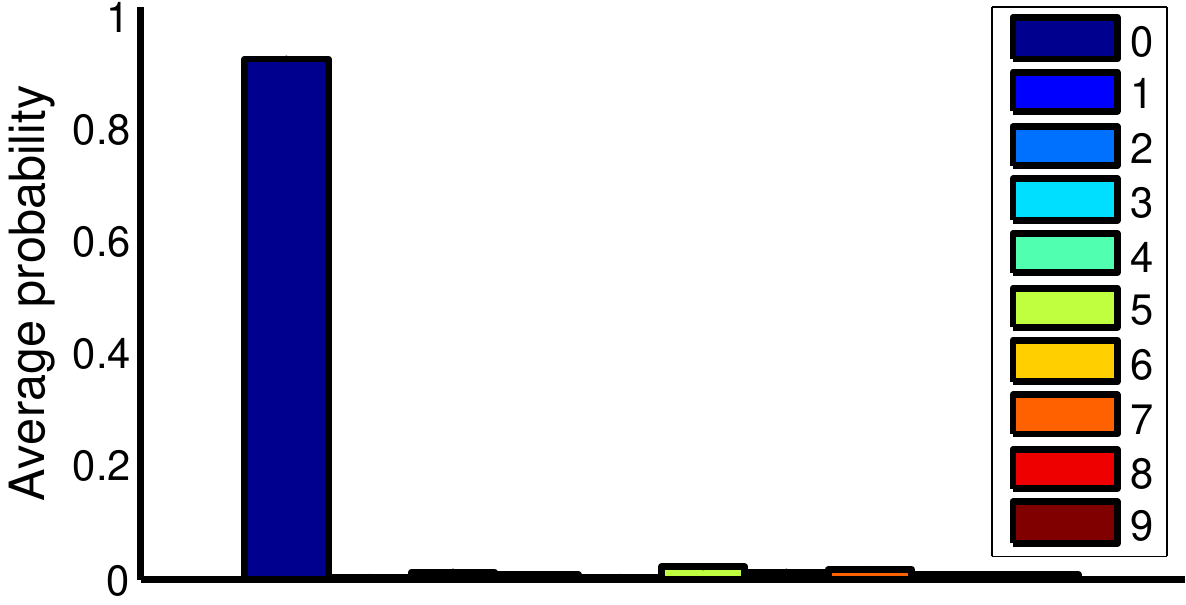}}} \\
        &&&&&&& \\
        &&&&&&& \\
        &&&&&&& \\
{}&
{\multirow{4}{*}{$k = 11$, $w_k = 7.2059$}} &
{\multirow{4}{*}{\includegraphics[height=.18\columnwidth]{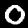}}} &
{\multirow{4}{*}{\includegraphics[height=.18\columnwidth]{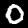}}} &
{\multirow{4}{*}{\includegraphics[height=.18\columnwidth]{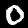}}} &
{\multirow{4}{*}{\includegraphics[height=.18\columnwidth]{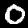}}} &
{\multirow{4}{*}{\includegraphics[height=.18\columnwidth]{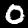}}} &
{\multirow{4}{*}{\includegraphics[height=.18\columnwidth]{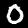}}} &
{\multirow{4}{*}{\includegraphics[width=.5\columnwidth,height=.19\columnwidth]{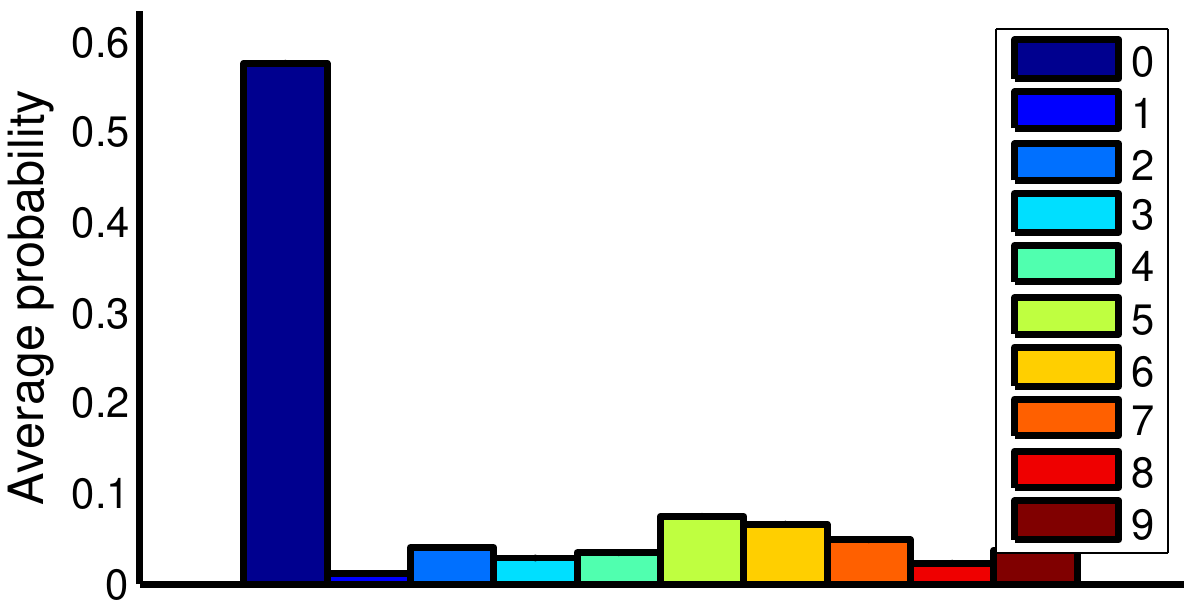}}} \\
        &&&&&&& \\
        &&&&&&& \\
        &&&&&&& \\
        \hline\hline
{\multirow{8}{*}{{\color{green} \bf neutral}}} &
{\multirow{4}{*}{$k = 55$,  $w_k = 0.4145$}} &
{\multirow{4}{*}{\includegraphics[height=.18\columnwidth]{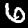}}} &
{\multirow{4}{*}{\includegraphics[height=.18\columnwidth]{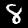}}} &
{\multirow{4}{*}{\includegraphics[height=.18\columnwidth]{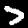}}} &
{\multirow{4}{*}{\includegraphics[height=.18\columnwidth]{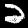}}} &
{\multirow{4}{*}{\includegraphics[height=.18\columnwidth]{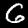}}} &
{\multirow{4}{*}{\includegraphics[height=.18\columnwidth]{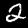}}} &
{\multirow{4}{*}{\includegraphics[width=.5\columnwidth,height=.19\columnwidth]{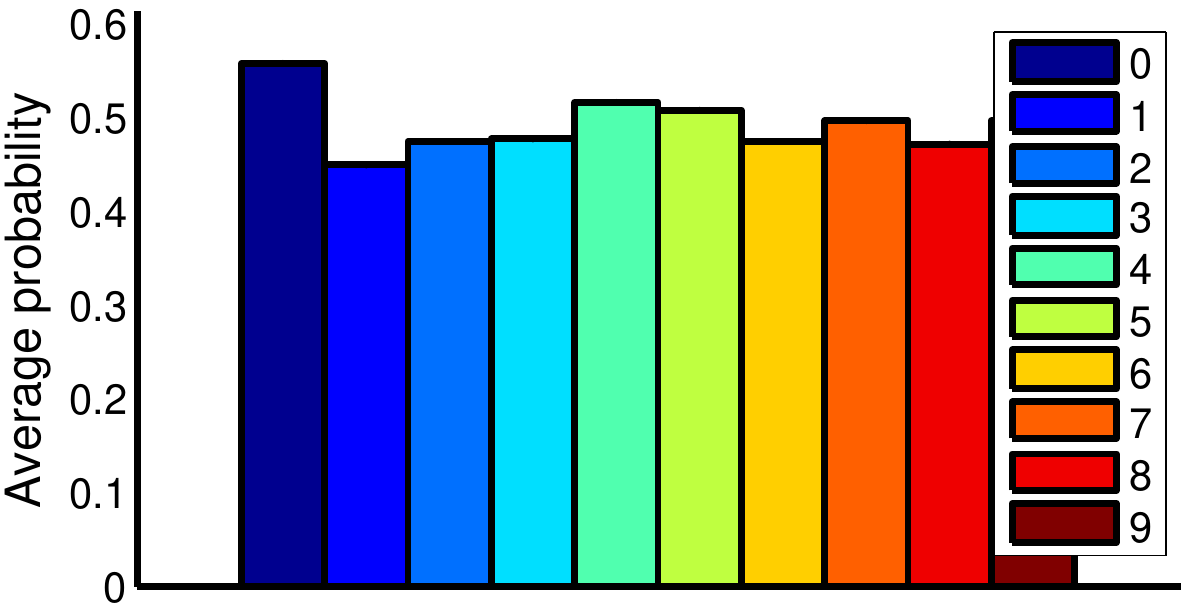}}} \\
        &&&&&&& \\
        &&&&&&& \\
        &&&&&&& \\
{}& {\multirow{4}{*}{$k = 59$, $w_k = -0.3355$}} &
{\multirow{4}{*}{\includegraphics[height=.18\columnwidth]{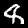}}} &
{\multirow{4}{*}{\includegraphics[height=.18\columnwidth]{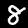}}} &
{\multirow{4}{*}{\includegraphics[height=.18\columnwidth]{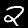}}} &
{\multirow{4}{*}{\includegraphics[height=.18\columnwidth]{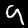}}} &
{\multirow{4}{*}{\includegraphics[height=.18\columnwidth]{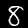}}} &
{\multirow{4}{*}{\includegraphics[height=.18\columnwidth]{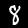}}} &
{\multirow{4}{*}{\includegraphics[width=.5\columnwidth,height=.19\columnwidth]{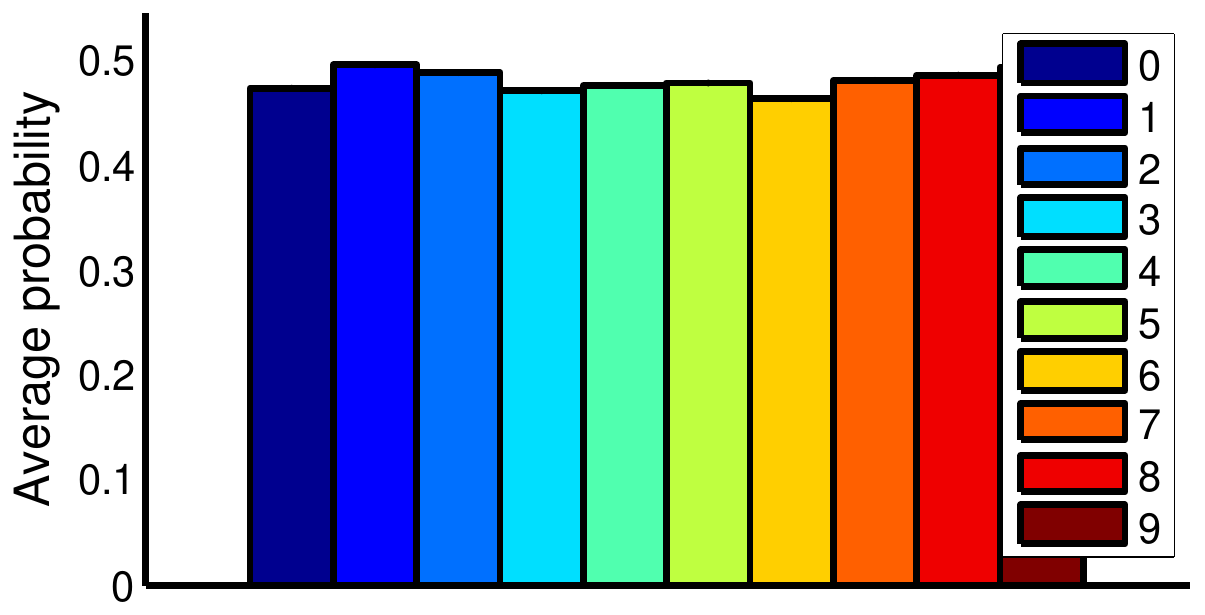}}} \\
        &&&&&&& \\
        &&&&&&& \\
        &&&&&&& \\
      \hline\hline
{\multirow{8}{*}{{\color{blue} \bf negative}}} &
{\multirow{4}{*}{$k = 13$, $w_k = -7.3152$}} &
{\multirow{4}{*}{\includegraphics[height=.18\columnwidth]{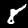}}} &
{\multirow{4}{*}{\includegraphics[height=.18\columnwidth]{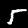}}} &
{\multirow{4}{*}{\includegraphics[height=.18\columnwidth]{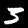}}} &
{\multirow{4}{*}{\includegraphics[height=.18\columnwidth]{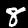}}} &
{\multirow{4}{*}{\includegraphics[height=.18\columnwidth]{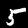}}} &
{\multirow{4}{*}{\includegraphics[height=.18\columnwidth]{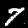}}} &
{\multirow{4}{*}{\includegraphics[width=.5\columnwidth,height=.19\columnwidth]{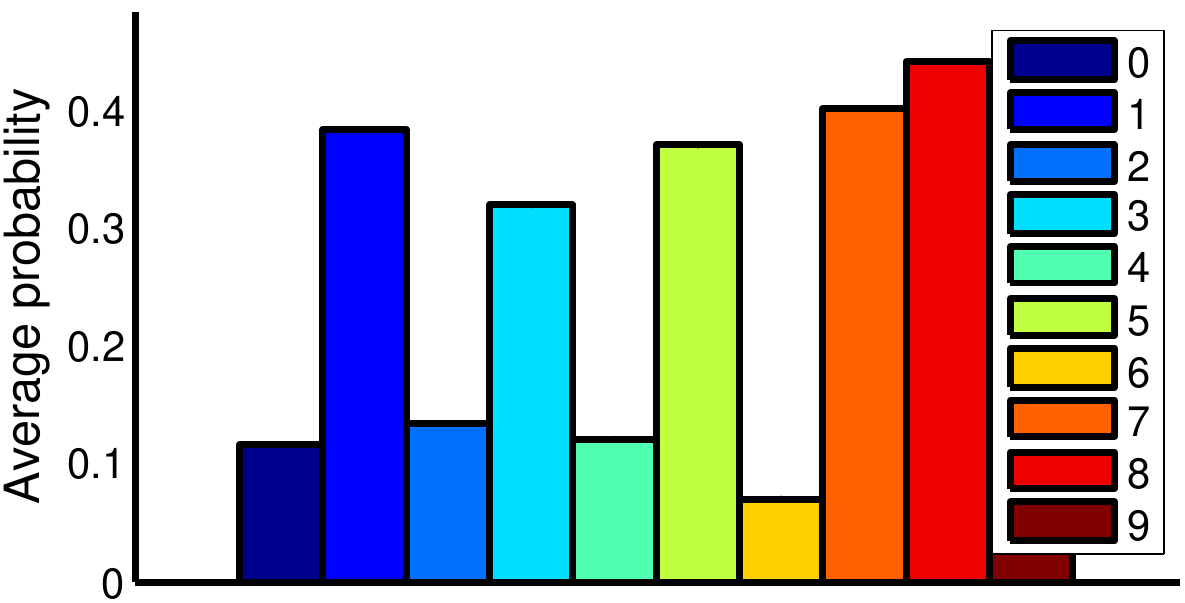}}} \\
        &&&&&&& \\
        &&&&&&& \\
        &&&&&&& \\
{}&{\multirow{4}{*}{$k = 5$, $w_k = -12.938$}} &
{\multirow{4}{*}{\includegraphics[height=.18\columnwidth]{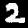}}} &
{\multirow{4}{*}{\includegraphics[height=.18\columnwidth]{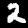}}} &
{\multirow{4}{*}{\includegraphics[height=.18\columnwidth]{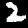}}} &
{\multirow{4}{*}{\includegraphics[height=.18\columnwidth]{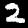}}} &
{\multirow{4}{*}{\includegraphics[height=.18\columnwidth]{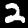}}} &
{\multirow{4}{*}{\includegraphics[height=.18\columnwidth]{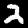}}} &
{\multirow{4}{*}{\includegraphics[width=.5\columnwidth,height=.19\columnwidth]{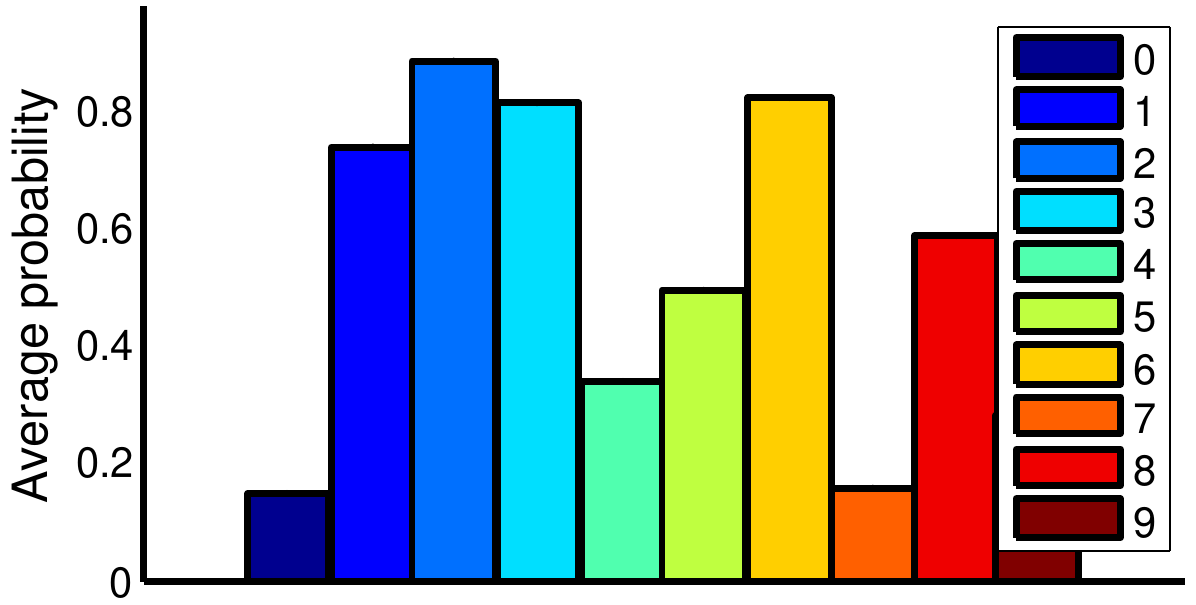}}} \\
        &&&&&&& \\
        &&&&&&& \\
        &&&&&&& \\
        \hline\hline
\end{tabular}}
\caption{(L) example images with highest responses to hidden units discovered by one of the binary classifiers (0 vs. others) of a 60 hidden-unit Dropout-LatentSVM model on the MNIST dataset. (R) average probabilities of each topic on representing images from the 10 categories.} \label{fig:mnist_visualize}
\end{figure*}

Fig.~\ref{NLCorruptSVM} shows the errors of nonlinear dropout classifiers compared with linear classifiers as a function of the random deletion percentage of features at the testing time. It can be observed that the nonlinear classifiers with latent representation significantly boost the prediction performance, which is consistent with the previous studies in the literature\footnote{http://yann.lecun.com/exdb/mnist/index.html}. Furthermore, by noting that Dropout-LatentSVM reduces to the standard LatentSVM with 1-layer perceptron, which is the case when the dropout level equals to zero, the dropout training strategy consistently boosts the performance when the deletion ratio is greater than $20\%$. Both Dropout-LatentSVM and Dropout-LatentLR are competitive for all types of feature deletion.

\begin{figure*}
 \centering
 \scalebox{.85}
 {
 \setlength{\tabcolsep}{3.9pt}
 \begin{tabular}{  c | c | c | c }
 \hline \hline
{\multirow{14}{*}{\includegraphics[height=.68\columnwidth]{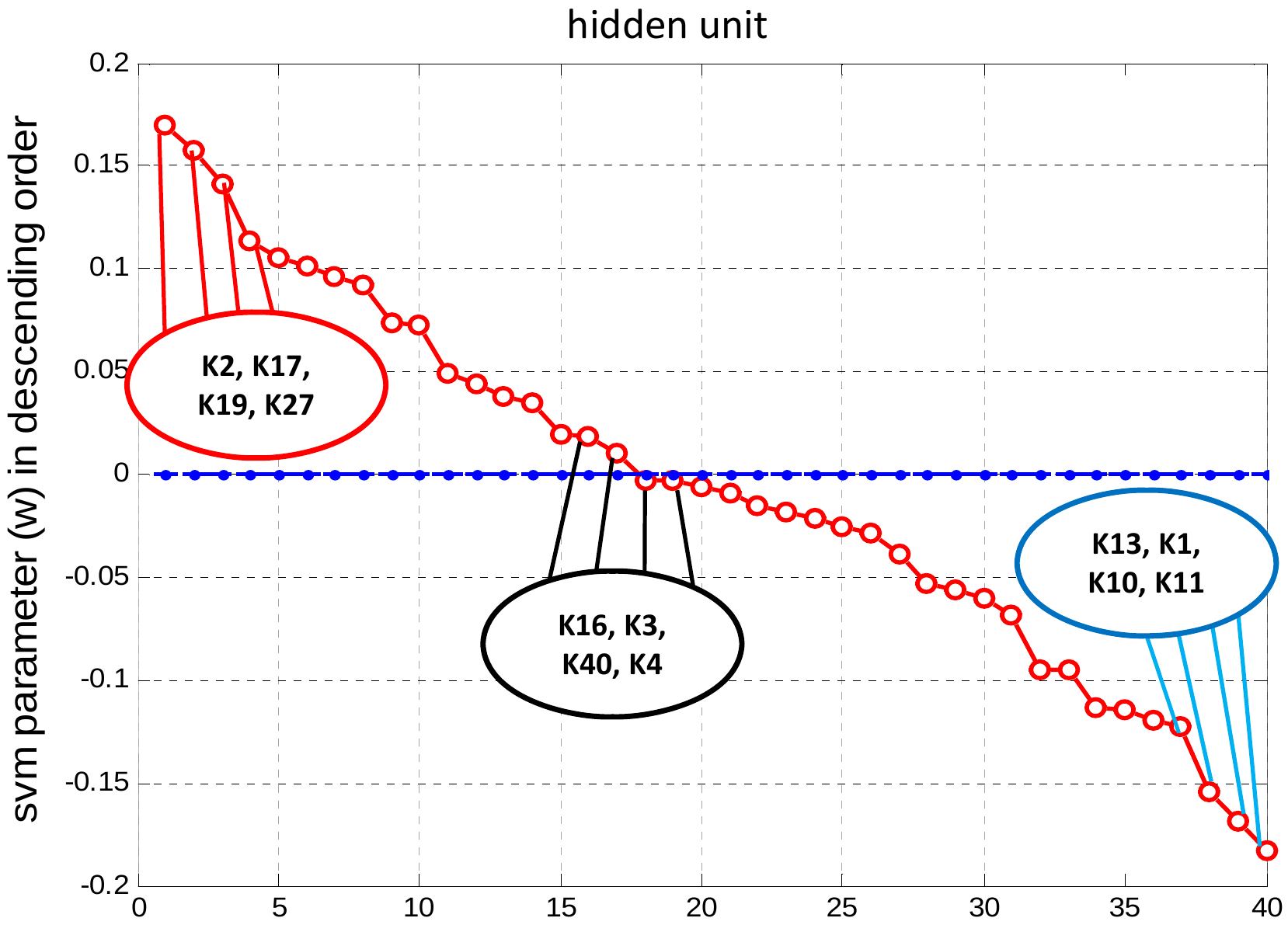}}} &
{\multirow{14}{*}{\includegraphics[height=.68\columnwidth]{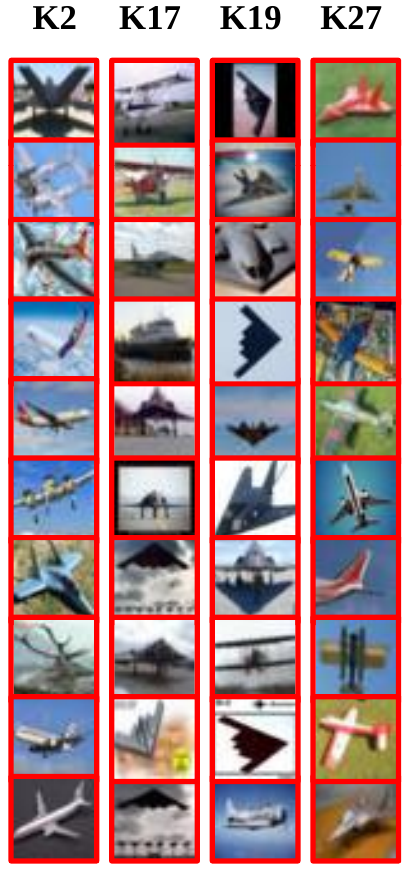}}} &
{\multirow{14}{*}{\includegraphics[height=.68\columnwidth]{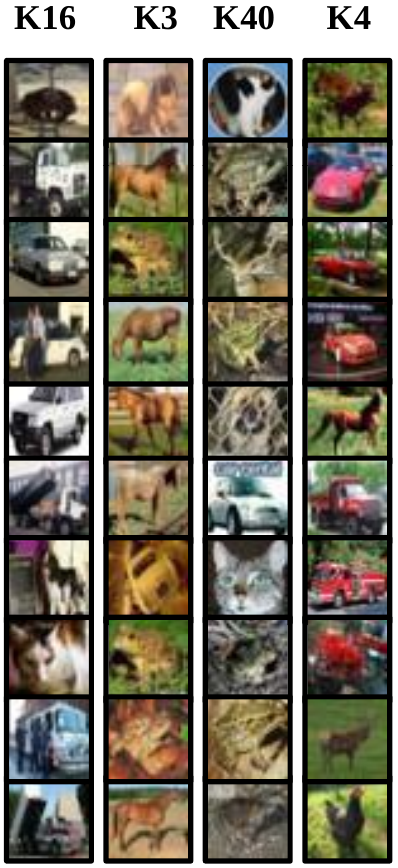}}} &
{\multirow{14}{*}{\includegraphics[height=.68\columnwidth]{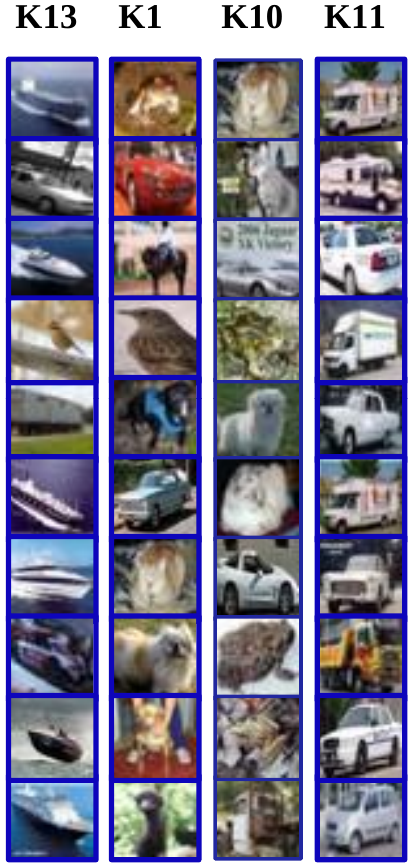}}} \\
        &&& \\
        &&& \\
        &&& \\
        &&& \\
        &&& \\
        &&& \\
        &&& \\
        &&& \\
        &&& \\
        &&& \\
        &&& \\
        &&& \\
        &&& \\
        \hline\hline
\end{tabular}}
\caption{Visualization on the CIFAR-10 dataset discovered by one of the binary classifiers (airplane vs. others) of a 40 hidden-unit Dropout-LatentSVM. (L) svm parameter values sorted in a descending order; (R) example images with highest responses to different hidden units that are associated with different svm parameter values.} \label{fig:cifar_visualize}\vspace{-.3cm}
\end{figure*}

\subsection{Latent Representation Visualization}

We examine various characteristics of the learned latent features by Dropout-LatentSVMs to show its ability in learning predictive latent subspace representations.

\subsubsection{MNIST}

We take a careful examination of each dimension in the discovered latent subspace of Dropout-LatentSVM. Note that our Dropout-LatentSVM is a binary classifier, and we use the ``one v.s. others" strategy for multi-class classification on the MNIST dataset. Here we choose the result of ``0 v.s. others" classifier to visualize the discriminative latent representations. Fig.~\ref{fig:mnist_visualize} shows six example hidden units (each unit corresponds to one dimension in the latent subspace) discovered by the Dropout-LatentSVM. For each hidden unit, we show six top-ranked images that yield higher expected value of $g_{\alphav}^k(\xv)$, together with the SVM parameter $w_k$. On the right side of Fig.~\ref{fig:mnist_visualize}, we show the average probability of each category distributed on the particular hidden unit. We can see that images with different $w_k$ values are very discriminative and predictive for several categories. For example, the first two hidden units ($k=2$ and $k=11$) with very positive $w_k$ values are discriminative in predicting the category ``0"; the last two hidden units ($k=55$ and $k=59$) with very negative $w_k$ values are good at discriminating a subset of categories $\{2, 3, 5, 7, 8\}$ against ``0"; while the middle two hidden units ($k=13$ and $k=5$) with $w_k$ values close to zero are kind of neutral and tend to represent all the categories.

\subsubsection{CIFAR-10}

Similarly, we examine the latent representations learned by Dropout-LatentSVM on the CIFAR-10 dataset. In Fig.~\ref{fig:cifar_visualize}, we take the ``airplane vs. others" binary classifier with 40 hidden units as an example to visualize the latent representations. On the left is the SVM parameter $w_k$ value sorted in descending order. The right side of Fig.~\ref{fig:cifar_visualize} shows the 3 groups of top-ranked example images of each latent dimension, where each group is associated with a different $w_k$ value (e.g., positive, neutral, negative), respectively. We can see that the latent subspace representations discovered by Dropout-LatentSVM is very expressive and discriminative. For example, the subgroup $\{2, 17, 19, 27\}$ with positive $w_k$ values is more representative of category ``airplane"; and the subgroup $\{13, 1, 10, 11\}$ with negative $w_k$ values tends to represent category \{``automobile", ``truck", ``ship", ``cat"\}; the subgroup $\{16, 3, 40, 4\}$ are neutral and tend to represent the categories \{``automobile", ``truck", ``bird", ``cat", ``horse", etc.\}.

\begin{figure*}
\centering
\subfigure[books]{\includegraphics[height=1.5in, width=1.6in]{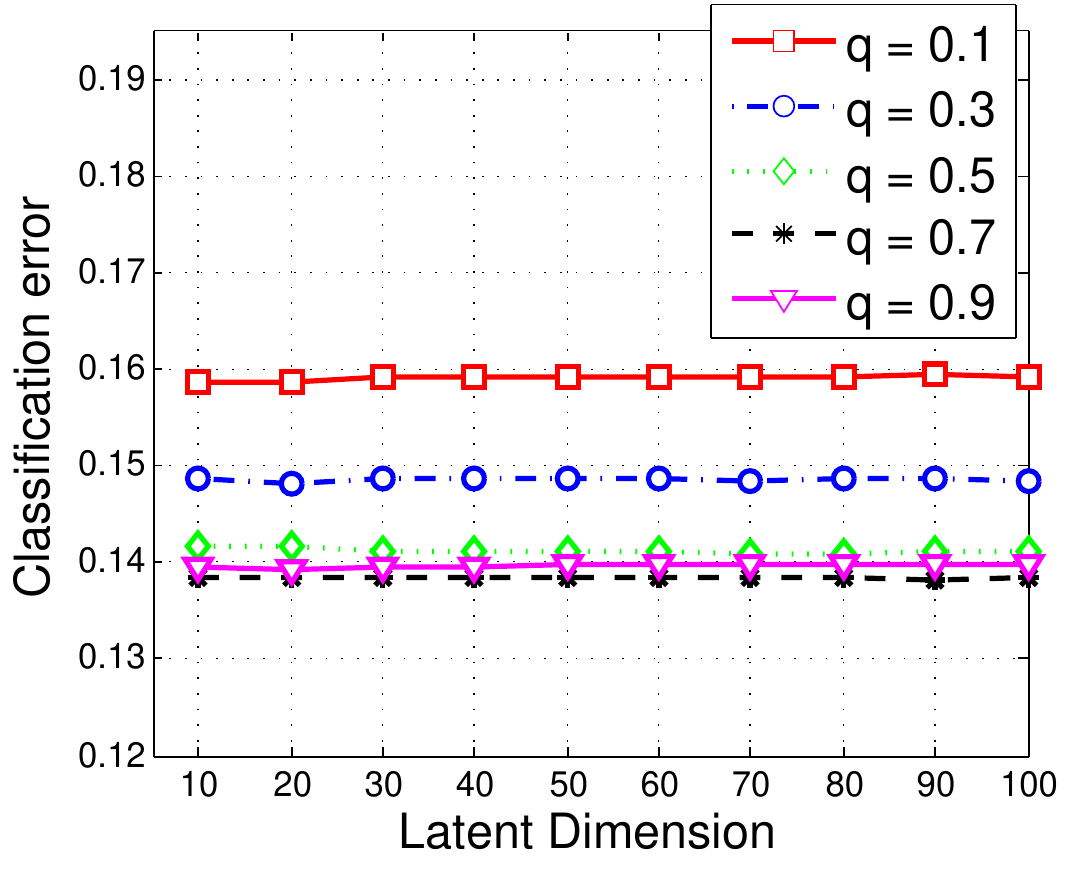}\label{sensK_books}}
\subfigure[kitchen]{\includegraphics[height=1.5in, width=1.6in]{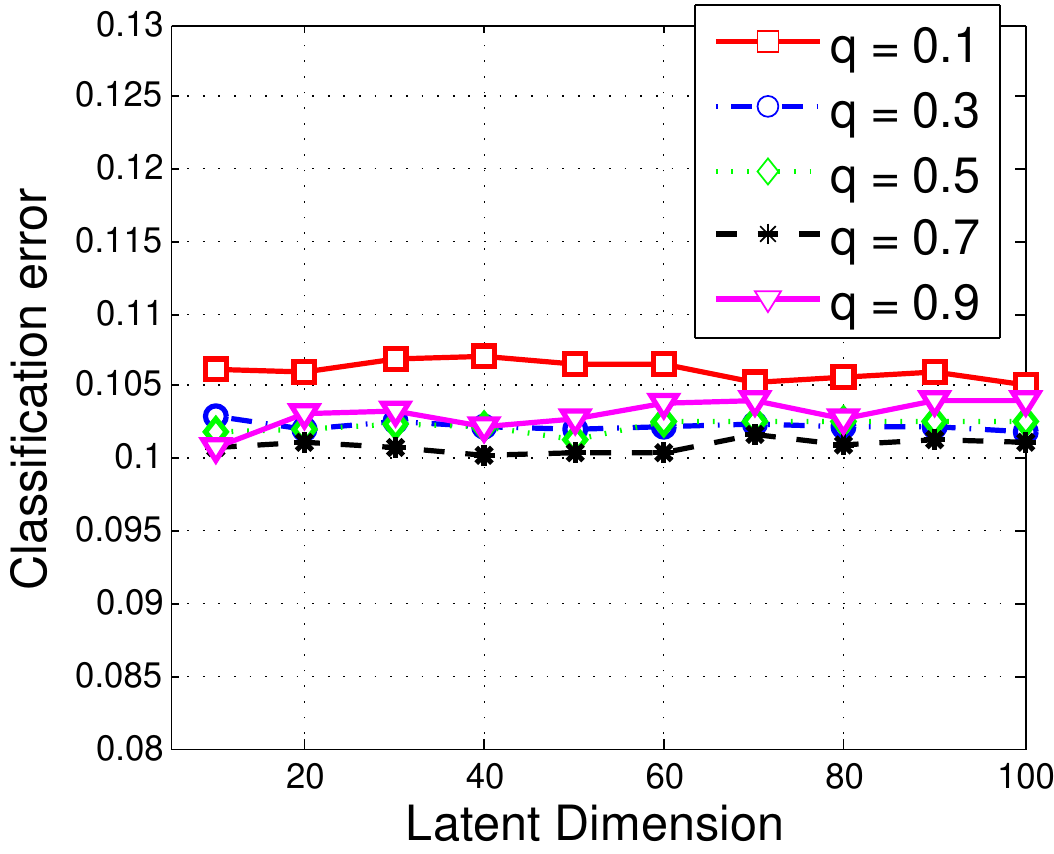}\label{sensK_kitchen}}
\subfigure[dvd]{\includegraphics[height=1.5in, width=1.6in]{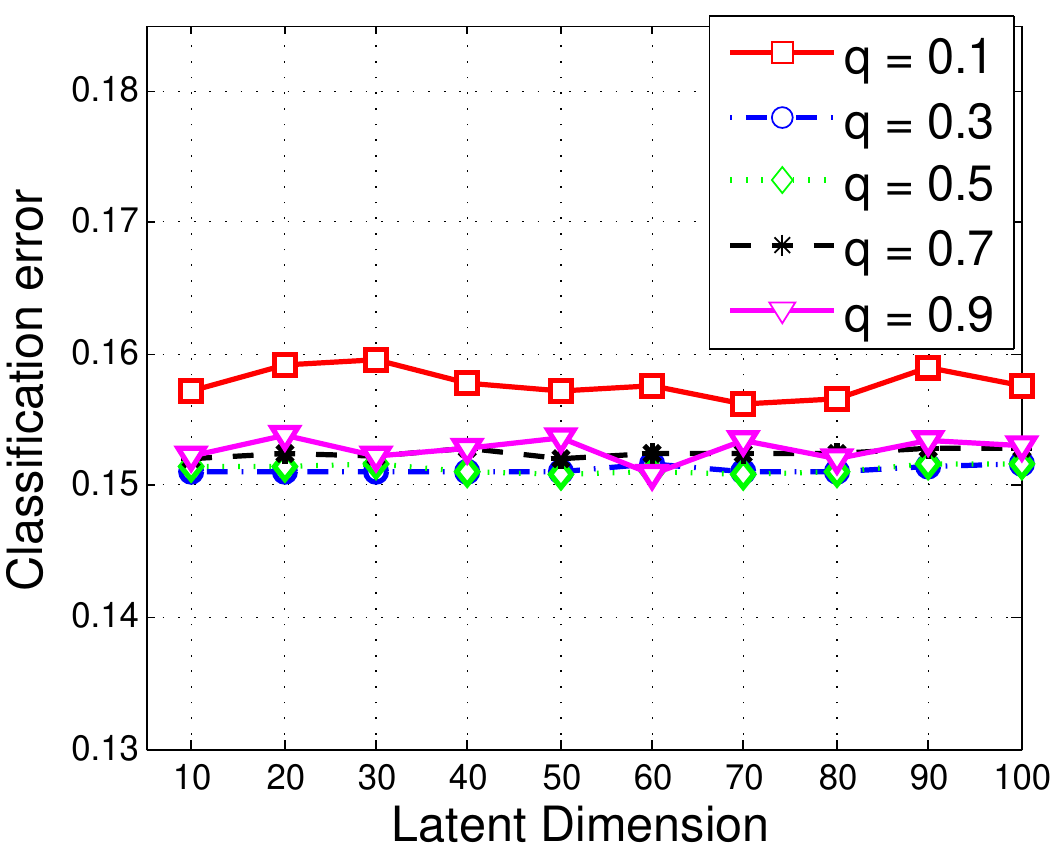}\label{sensK_dvd}}
\subfigure[electronics]{\includegraphics[height=1.5in, width=1.6in]{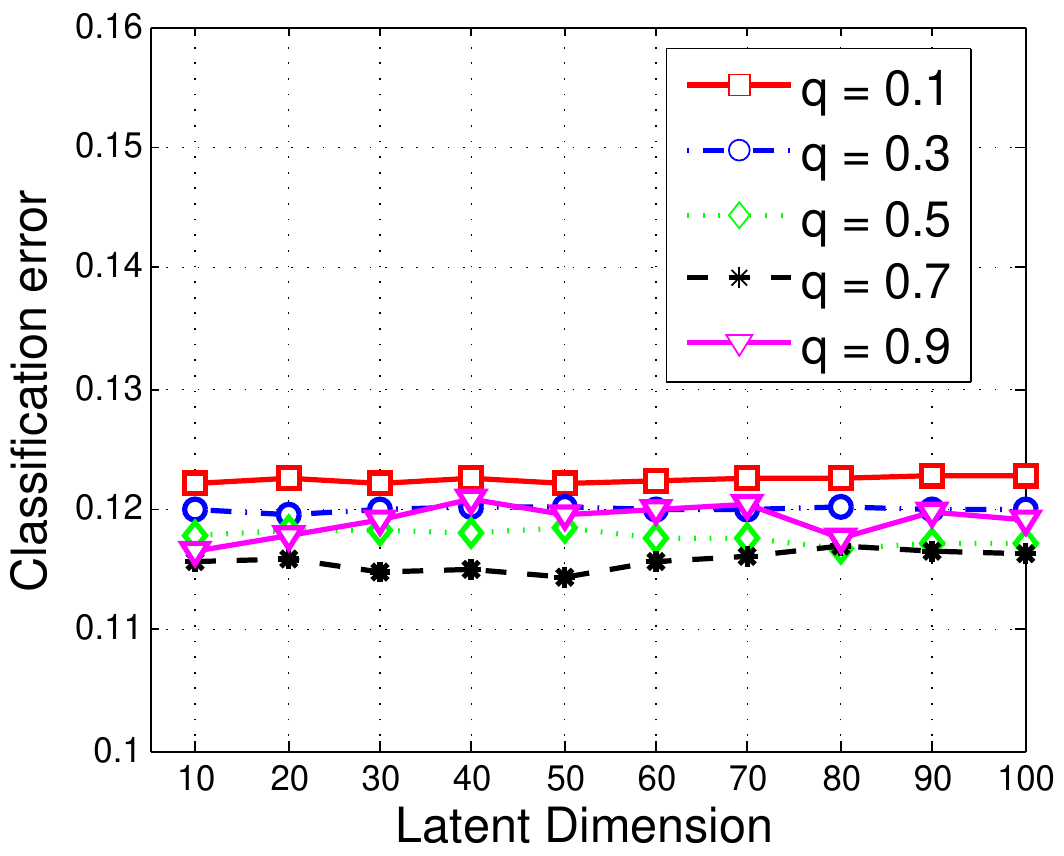}\label{sensK_electronics}}\vspace{-.2cm}
\caption{Sensitivity of Dropout-LatentSVM to Latent Dimension on Amazon review datasets.}
\label{fig:LatentDim1}\vspace{-.3cm}
\end{figure*}

\subsection{Sensitivity to Latent Dimension}

To provide more insights about the behavior of Dropout classifiers, 
we investigate the prediction performance of nonlinear classifiers with respect to the latent dimensions. Fig.~\ref{fig:LatentDim1} shows the classification errors of Dropout-LatentSVM on four Amazon review datasets with different latent dimensions. We can see that Dropout-LatentSVM using different dropout levels are insensitive to the latent dimensions on all datasets. We have similar observations for Dropout-LatentLR. 

\subsection{Time Complexity}

\begin{figure}\vspace{-.2cm}
\centering
\subfigure[Train Time]{\includegraphics[height=1.1in]{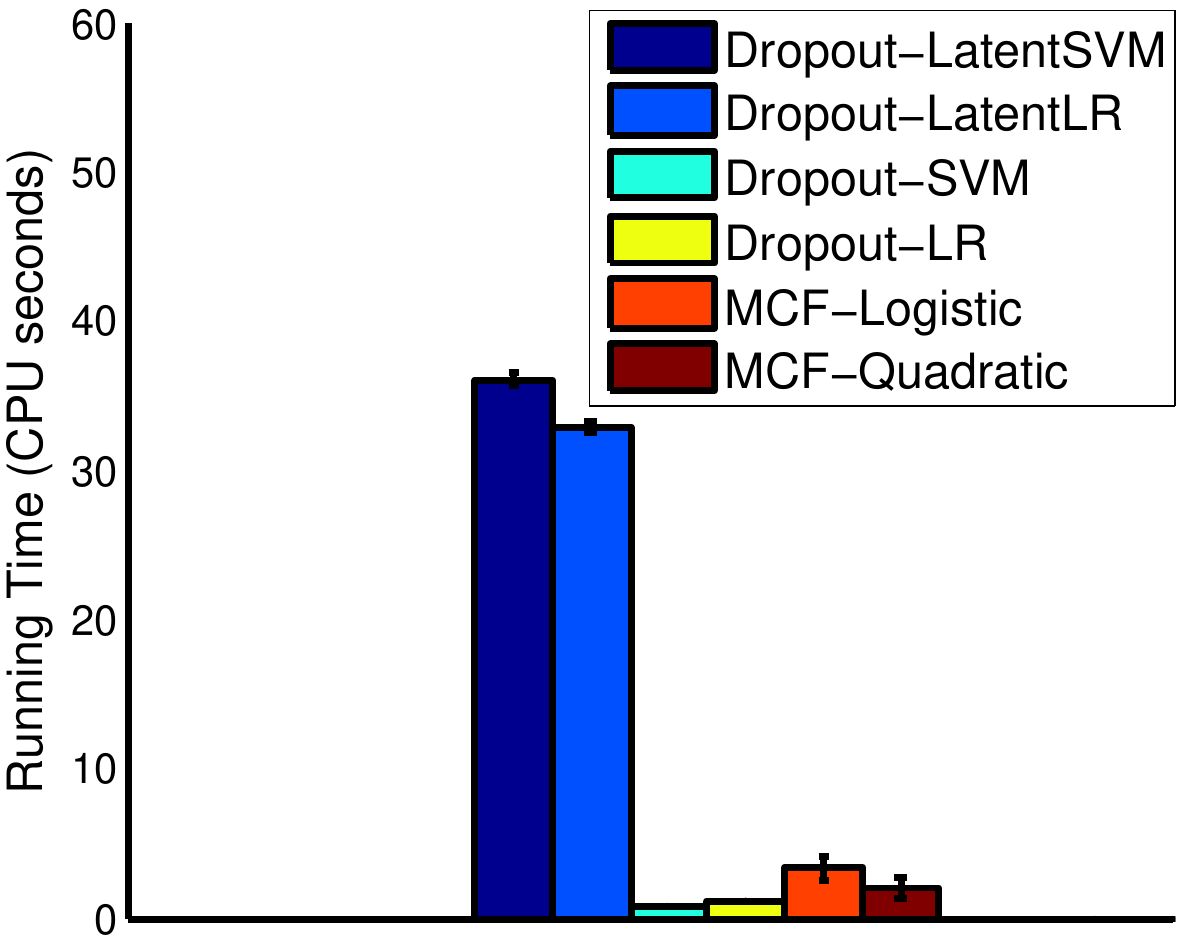}\label{TrTime}}
\subfigure[Test Time]{\includegraphics[height=1.1in]{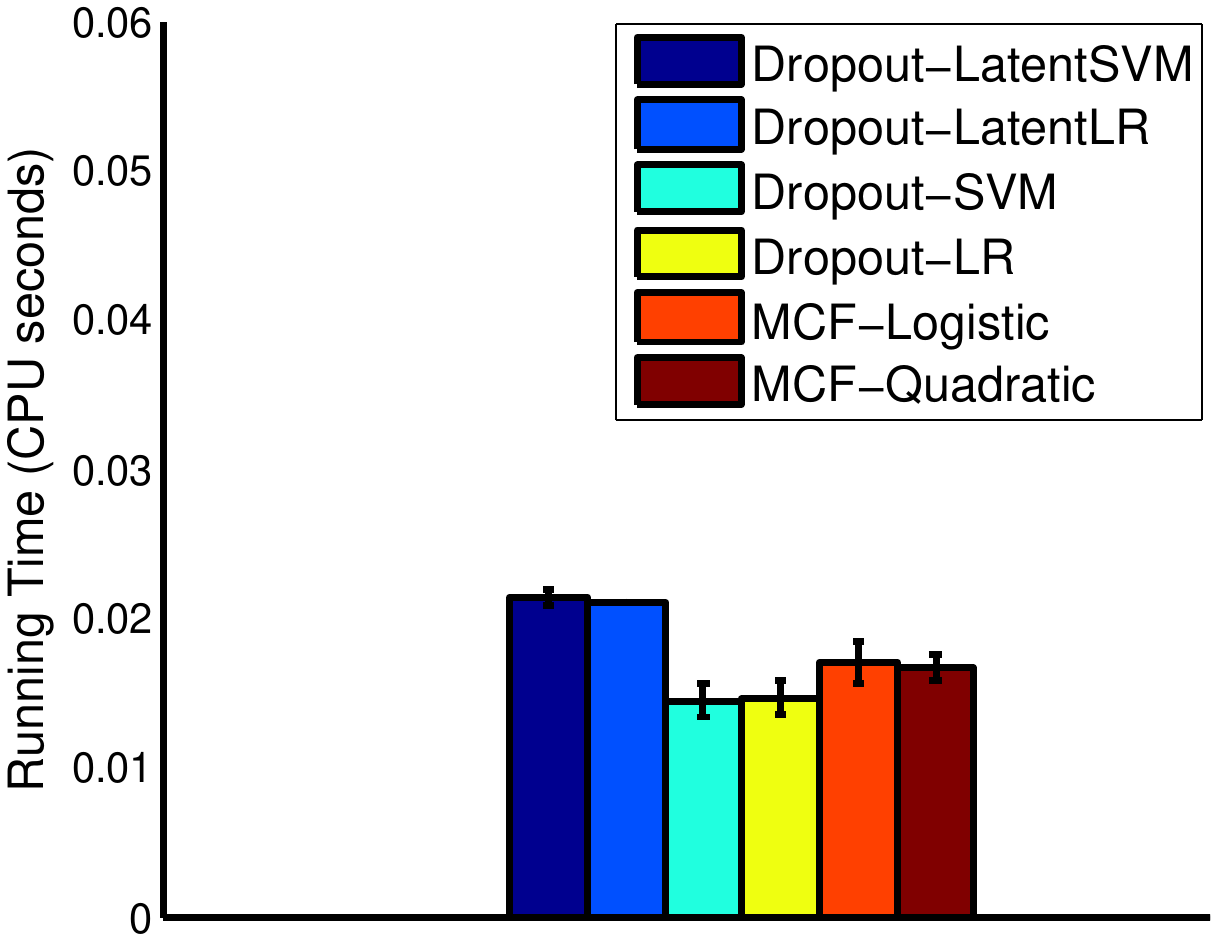}\label{TstTime}}\vspace{-.2cm}
\caption{Time comparison of different models on amazon books dataset.}
\label{fig:time}\vspace{-.4cm}
\end{figure}

Fig.~\ref{fig:time} compares the time efficiency of both linear and nonlinear Dropout-SVM, Dropout-LR models with MCF-Logistic and MCF-Quadratic models on the Amazon-books review dataset. The four proposed models (i.e., Dropout-SVM, Dropout-LR, Dropout-LatentSVM and Dropout-LatentLR) are implemented in C++, and we use the matlab implementation of MCF-Logistic and MCF-Quadratic. All the models are run on a 3.40GHz desktop with 4GB RAM. For training, we can observe that: 1) the time cost of linear classifiers (i.e., Dropout-SVM and Dropout-LR) are comparable with (slightly faster than) MCF-Logistic and MCF-Quadratic models, which shows the efficiency of our proposed methods; 2) Dropout-SVM and Dropout-LR models are more efficient than the Dropout-LatentSVM and Dropout-LatentLR models, which is reasonable as nonlinear classifiers need to learn one-hidden layer perceptron. For testing, all the models are deterministic and very efficient for making predictions.

\section{Conclusions and Future Work}\label{sec:conclusion}

We present dropout training for both linear SVMs and its nonlinear extension by learning latent features, with an iteratively re-weighted least square (IRLS) algorithm by using data augmentation techniques. Similar ideas are applied to develop a new IRLS algorithm for the dropout training of logistic regression. Our IRLS algorithms provide insights on the connection and difference among various losses in dropout learning settings. Empirical results on various tasks demonstrate the effectiveness of our approaches.

For future work, it is remained open whether the kernel trick can be incorporated in dropout learning. We are also interested in developing more efficient algorithms, e.g., online dropout learning, to deal with even larger datasets, and investigating whether Dropout-SVM can be incorporated into a deep learning architecture~\cite{Tang:2013} or learning with latent structures~\cite{Zhu:jmlr14} and in the context of hierarchical Bayes networks~\cite{MaxWelling:2014}. We are also interested in designning better and more informed dropout policies, e.g., using reinforcement learning techniques~\cite{SelectDropoutLevel:2015}.




\ifCLASSOPTIONcompsoc
  \section*{Acknowledgments}
\else
  \section*{Acknowledgment}
\fi

This work is supported by National Key Project for Basic Research of China (Nos: 2013CB329403, 2012CB316301), National Natural Science Foundation of China (Nos: 61305066, 61322308, 61332007), Tsinghua Self-innovation Project (No: 20121088071) and China Postdoctoral Science Foundation Grant (Nos: 2013T60117, 2012M520281).

\ifCLASSOPTIONcaptionsoff
  \newpage
\fi
{\small
\bibliographystyle{plain}
\bibliography{DropoutSVM_v9}
}
\begin{IEEEbiography}[{\includegraphics[width=.9in,height=1.15in]{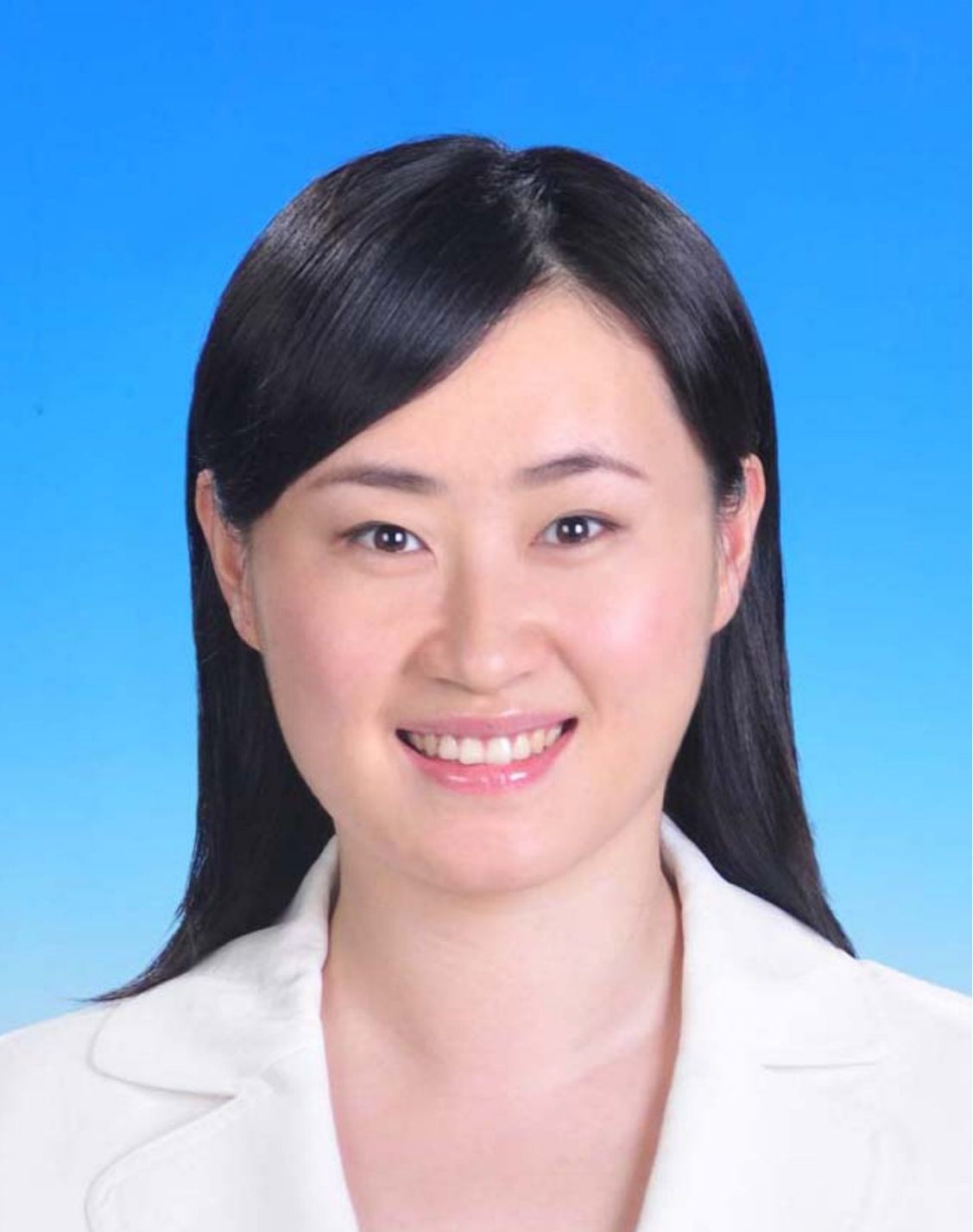}}]
{Ning Chen}
received her PhD degree in the Department of Computer Science and Technology at Tsinghua University, China, where she is currently an assistant researcher. She was a visiting researcher in the Machine Learning Department of Carnegie Mellon University. Her research interests are primarily in machine learning, especially probabilistic graphical models with applications on data mining and bioinformatics.
\end{IEEEbiography}

\begin{IEEEbiography}[{\includegraphics[width=.9in,height=1.1in]{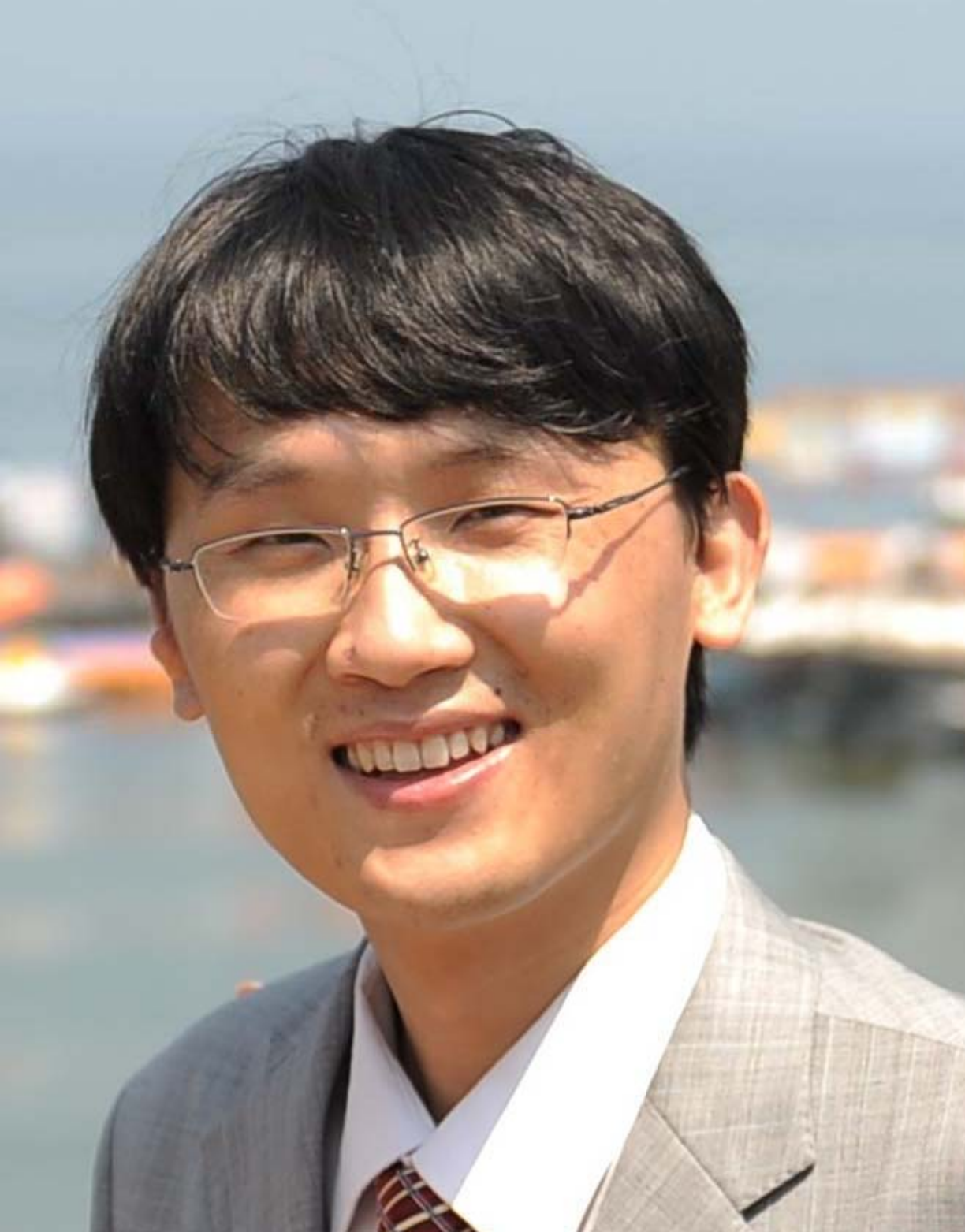}}]
{Jun Zhu}
received his BS, MS and PhD degrees all from the Department of Computer Science and Technology at Tsinghua University, China, where he is currently an associate professor. He was a project scientist and postdoctoral fellow in the Machine Learning Department, Carnegie Mellon University. His research interests are on developing machine learning methods to understand scientific/engineering data arising from various fields. He is a member of the IEEE.
\end{IEEEbiography}

\begin{IEEEbiography}[{\includegraphics[width=.9in,height=1.1in]{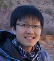}}]
{Jianfei Chen}
received his BS from Department of Computer Science and Technology, Tsinghua University, China, where he is currently a Phd student. His research interests are primarily on machine learning especially on probabilistic graphical models, Bayesian nonparametrics and data mining problems such as social networks.
\end{IEEEbiography}


\begin{IEEEbiography}[{\includegraphics[width=.9in,height=1.15in]{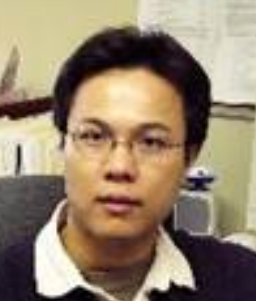}}]
{Ting Chen} received his BS in Computer Science from Tsinghua University in 1993 and PhD degree from SUNY Stony Brook in 1997. He is currently a professor in Tsinghua National Lab for Information Science and Technology. He was a professor of biological sciences, computer science, and mathematics at University of Southern California. His research interests are in applying machine learning and computer algorithms to answer questions in biology and medicine. 
\end{IEEEbiography}





\section*{Appendix A: Variational Upper Bound}\label{sec:appendix}

We provide details on deriving the variational bound of the expected hinge loss in (\ref{eqn:EpHingeLoss}). 
To simplify notations, we derive the bound for a single data point. For a dataset with $N$ examples, a simple summation will give the final bound. Define $g(\thetav; \xv) \triangleq \ep_p \lbrack \log \phi(y | \tilde{\xv}, \thetav) \rbrack$. We have
\begin{eqnarray}
g(\thetav; \xv) 
&=& \ep_{p} \Big\lbrack \log \int \frac{1}{\sqrt{2\pi \lambda}} \exp\Big\{-\frac{(\lambda +c\zeta ))^2}{2\lambda} \Big\} d\lambda \Big\rbrack\nonumber\\
&=& \ep_{p} \Big\lbrack \log \int \frac{q(\lambda)}{q(\lambda)\sqrt{2\pi\lambda}}\exp\Big\{-\frac{(\lambda +c\zeta)^2}{2\lambda}\Big\} d\lambda \Big\rbrack\nonumber\\
&\geq& \ep_{p}\Big\lbrack \ep_{q(\lambda)} \log\frac{1}{q(\lambda)\sqrt{2\pi\lambda}} \exp\Big\{-\frac{(\lambda + c\zeta)^2}{2\lambda}\Big\}\Big\rbrack\nonumber\\
&=& \Big\{ H(\lambda) - \frac{1}{2}\ep_q \lbrack\log\lambda\rbrack - \ep_q \Big\lbrack
\frac{1}{2\lambda} \ep_{p} (\lambda + c\zeta)^2\Big\rbrack\Big\} + c^\prime \nonumber
\end{eqnarray}
where $\lambda$ is the augmented variable and $c^\prime$ is a constant. Note that
if there is no uncertainty in the feature corruption (e.g., the corruption level in the dropout or blankout noise is 0), the bound is tight. That is, the optimal solution of $q$ will give the original hinge loss.


\section*{Appendix B: Proof of Lemma 1}\label{sec:appendixB}
\begin{proof}
Ignore the $\ell_2$-norm regularizer, we have the objective of the M-step:
\setlength\arraycolsep{-2pt} \begin{eqnarray}
&& \mathcal{L}_{[\wv]}  =  \sum_{n=1}^N  \ep_{p}\left[c \zeta_n  +  \frac{c^2}{2} \gamma_n \zeta_n^2 \right],
\end{eqnarray}
where $\gamma_n \triangleq \ep_q[\lambda_n^{-1}]$. Using the definition of $\zeta_n \triangleq \ell - y_n \wv^\top \tilde{\xv}_n$ and ignoring the constants, we have the simplified objective function (again without the $\ell_2$-regularizer):
\setlength\arraycolsep{1pt} \begin{eqnarray}
 \mathcal{L}_{[\wv]} && = \sum_{n=1}^N \ep_p \left[   \frac{c^2}{2} \gamma_n \wv^\top  \tilde{\xv}_n \tilde{\xv}_n^\top  \wv  - (c + \ell c^2 \gamma_n ) y_n \wv^\top  \tilde{\xv}_n  \right] \nonumber \\
&& = \frac{c^2}{2} \sum_{n=1}^N \gamma_n \ep_p \left[ \wv^\top \tilde{\xv}_n \tilde{\xv}_n^\top  \wv  - 2y_n^h \wv^\top  \tilde{\xv}_n \right] \nonumber \\
&& = \frac{c^2}{2} \sum_{n=1}^N \gamma_n \ep_p \left[ (\wv^\top \tilde{\xv}_n - y_n^h)^2 \right],
\end{eqnarray}
where $y_n^h \triangleq ( \frac{1}{c \gamma_n} + \ell) y_n$ is the re-weighted label.

We now derive the equations to compute $\gamma_n$.
Let $x = \lambda_n$, and $f(x) = \frac{1}{\lambda_n}$. By the transformation rule of probability distributions that $p(x)=p(f(x))|\frac{d f(x)}{dx}|$, we have $q(\lambda_n)=\frac{1}{\lambda_n^2}q(\frac{1}{\lambda_n})$. Then
\begin{eqnarray}
\ep_{q(\lambda_n)} [\lambda_n^{-1}] && = \int_{0}^{\infty} q(\lambda_n) \frac{1}{\lambda_n}d\lambda_n \nonumber \\
                    && =\int_{\infty}^0 q(\mu_n) \mu_n^3 d \mu_n^{-1}~~(\textrm{define}~\mu_n = \frac{1}{\lambda_n}) \nonumber \\
                    && = \int_0^{\infty} q(\mu_n)\mu_n d\mu_n \nonumber \\
                    && = \frac{1}{c \sqrt{\ep[\zeta_n^2]}},
\end{eqnarray}
where the last equality is due to the fact that $q(\lambda_n^{-1})$ or equivalently $q(\mu_n)$ is an inverse Gaussian distribution as shown in Eq. (\ref{eqn:inverseGaussian}).
\end{proof}

\end{document}